\documentclass[preprint,12pt,authoryear]{elsarticle}

\usepackage[left]{lineno}

\usepackage{rotating} 
\usepackage{graphicx}

\usepackage{eso-pic}
\usepackage[utf8]{inputenc}
\usepackage{microtype}
\usepackage{float}
\usepackage[margin=1in]{geometry} 
\usepackage{multicol}
\usepackage[hidelinks, colorlinks=true,linkcolor=blue]{hyperref}
\usepackage[section]{placeins}
\usepackage{amsmath}
\usepackage{microtype}
\usepackage{url}     
\usepackage{setspace}
\usepackage{longtable}

\usepackage[T1]{fontenc}
\usepackage{lmodern}
\usepackage{amsmath,amssymb}
\usepackage{graphicx}
\usepackage{caption}
\usepackage{subcaption}
\usepackage[hidelinks]{hyperref}
\usepackage{lineno}
\usepackage[authoryear]{natbib}
\usepackage{caption,tabularx,booktabs}
\usepackage{siunitx}
\usepackage{float}
\usepackage{graphicx}
\usepackage{amssymb}
\usepackage{tabularx}
\usepackage{amsmath}
\usepackage{amsfonts}

\begin{document}
\makeatletter
\def\ps@pprintTitle{%
  \let\@oddhead\@empty
  \let\@evenhead\@empty
  \let\@oddfoot\@empty
  \let\@evenfoot\@oddfoot
}
\makeatother

\title{Interpretable Machine Learning for Reservoir Water Temperatures in the U.S. Red River Basin of the South}


%
\author[1]{Isabela Suaza-Sierra}

\ead{isuazasier@miners.utep.edu}

\affiliation[1]{organization= {Department of Earth, Environmental and Resource Sciences, The University of Texas at El Paso},
    city={El Paso},
    state={TX},
    postcode={79968}, 
    country={USA}}

\author[1]{Hernan A. Moreno}

\author[1]{Luis A De la Fuente}

\author[2]{Thomas M. Neeson}

\affiliation[2]{organization={Department of Geography and Environmental Sustainability, University of Oklahoma},
    city={Norman},
    state={OK},
    postcode={73019}, 
    country={USA}}

\begin{abstract}

Accurate prediction of Reservoir Water Temperature (RWT) is vital for sustainable water management, ecosystem health, and climate resilience. Yet, prediction alone offers limited insight into the governing physical processes. To bridge this gap, we integrated explainable machine learning (ML) with symbolic modeling to uncover the drivers of RWT dynamics across ten reservoirs in the Red River Basin, USA, using over 10,000 depth-resolved temperature profiles.

We first employed ensemble and neural models, including Random Forest (RF), Extreme Gradient Boosting (XGBoost), and Multilayer Perceptron (MLP),achieving high predictive skill (best RMSE = 1.20 °C, R² = 0.97). Using SHAP (SHapley Additive exPlanations), we quantified the contribution of physical drivers such as air temperature, depth, wind, and lake volume, revealing consistent patterns across reservoirs.

To translate these data-driven insights into compact analytical expressions, we developed Kolmogorov–Arnold Networks (KANs) to symbolically approximate RWT. Ten progressively complex KAN equations were derived, improving from R² = 0.84 using a single predictor (7-day antecedent air temperature) to R² = 0.92 with ten predictors, though gains diminished beyond five, highlighting a balance between simplicity and accuracy.

The resulting equations, dominated by linear and rational forms, incrementally captured nonlinear behavior while preserving interpretability. Depth consistently emerged as a secondary but critical predictor, whereas precipitation had limited effect. By coupling predictive accuracy with explanatory power, this framework demonstrates how KANs and explainable ML can transform black-box models into transparent surrogates that advance both prediction and understanding of reservoir thermal dynamics.



\end{abstract}

\maketitle 
\section{Introduction}
Reservoir Water Temperature (RWT) is a fundamental variable in freshwater systems, shaping aquatic organisms’ physiology, behavior, and spatial distribution \citep{caissie2006thermal}. It regulates key ecosystem functions including thermal stratification, nutrient cycling, and dissolved oxygen dynamics \citep{adrian2009lakes, davis2024effects}. Operationally, RWT serves as a critical input for reservoir management, guiding decisions in hydropower generation, water supply, and flood control \citep{magee2017response, imberger1989physical}. Warmer water temperatures can elevate the risk of hypoxia and harmful algal blooms, exacerbating water quality concerns \citep{johnk2008summer}. Furthermore, RWT is a sensitive indicator of climate variability and change, responding to changes in atmospheric temperature, wind, and precipitation patterns \citep{song2016spatiotemporal, schmid2014lake}. Modeling RWT thus enhances our capacity to predict and adapt to climate-driven hydrological changes.

Water temperature in the epilimnion (the upper, well-mixed layer of a stratified lake) responds rapidly to atmospheric forcing, making it a reliable indicator of short-term hydrometeorological trends \citep{robertson1990changes, Piccolroaz}. In contrast, hypolimnetic temperatures (those in the cold, dense bottom layer isolated from surface interactions) reflect more complex dynamics shaped by lake morphometry \citep{gerten2000climate}, seasonality, and longer-term climate signals \citep{robertson1990changes}. The vertical distribution of temperature governs density gradients, which in turn influence vertical mixing, stratification regimes, and thermocline depth \citep{woolway2022lakes,coats2006warming}. Seasonal stratification regulates nutrient redistribution and biological productivity \citep{liu2018thermal}, while physicochemical gradients can directly affect fish health and ecosystem stability \citep{jeppesen2012impacts}.

Reservoir water temperature is shaped by a complex interplay of meteorological, hydrological, and morphometric drivers, each exerting influence across distinct temporal and spatial scales. Among these, air temperature is consistently identified as the dominant atmospheric driver, controlling surface energy exchanges through sensible heat flux and dictating the net atmospheric heat input to surface waters \citep{robertson1990changes,liu2018thermal}. Diurnal and seasonal variability in air temperature contributes directly to surface water warming rates and the persistence of thermal stratification \citep{woolway2016diel}.

Solar radiation, particularly shortwave radiation, penetrates the water column and is absorbed at different depths based on turbidity, colored dissolved organic matter, and particulate content, driving heating in the epilimnion \citep{winslow2017seasonality,piccolroaz2013simple, woolway2017amplified}. In contrast, longwave radiation and evaporative cooling govern nighttime and seasonal cooling cycles, influencing energy loss from the surface and contributing to vertical heat redistribution \citep{austin2011sensitivity,stepanenko2013one}.

Wind speed and direction affect water temperature indirectly but powerfully by driving surface shear stress and vertical turbulent mixing. Wind-induced mixing can deepen the mixed layer and disrupt stratification or cause full overturn, especially in shallower or polymictic systems \citep{imberger1989physical, magee2017response}. In deeper lakes and reservoirs, such mixing is often confined to upper layers, leaving hypolimnetic temperatures less influenced by daily wind dynamics \citep{stefan2001simulated}.

Hydrological inflows and outflows modulate reservoir temperature profiles through thermal density currents and advective heat transport. Cold snowmelt or warm runoff entering at different times of year can intrude at intermediate depths, creating temporary stratification or thermal barriers within the metalimnion \citep{goudsmit2002application}. Outflows can remove specific water layers depending on withdrawal elevation, altering temperature distributions and residence time \citep{casamitjana2003effects}.

Lastly, reservoir morphometry, including surface area, depth, and hypsographic shape, mediates all these processes by influencing heat storage capacity and stratification stability \citep{kirillin2016generalized}. These physical drivers act interactively and often nonlinearly, resulting in complex and site-specific thermal dynamics. Their influence varies not only between reservoirs but also with time, particularly under changing climate conditions and reservoir regulation schemes. Understanding their mechanisms is essential for accurately modeling water temperature, anticipating ecological consequences, and informing management decisions in the context of warming and hydrological change \citep{wang2023climate, woolway2021phenological}.\\
Traditional RWT models, both physics-based and empirical, simulate complex interactions involving heat fluxes (radiation, evaporation, conduction), mass exchanges (inflows, outflows, vapor), and mechanical forces such as wind-driven mixing \citep{woolway2021phenological}. These models often simulate vertical mixing and stratification influenced by temperature, chemistry, and biology \citep{tanentzap2007}, but suffer from limited scalability, site dependency, and heavy computational demands \citep{munz2017fluxbot}. Empirical models offer computational efficiency but typically oversimplify the system, failing to capture nonstationary and nonlinear behavior 
\citep{rashid2019simulation}. Even widely used models like the General Lake Model (GLM) require extensive calibration and are sensitive to input uncertainty under climate scenarios 
\citep{fenocchi2018forecasting,abbasian2020multi}.

ML approaches, including Random Forest (RF) \citep{breiman2001random}, Extreme Gradient Boosting (XGBoost) \citep{XGBoost}, and Multilayer Perceptron (MLP) \citep{rumelhart1986learning} have gained traction for their ability to learn from large datasets and generalize across hydrological systems. Examples include RF applications in evaporation forecasting from reservoirs in Ethiopia \citep{eshetu2023interpretable} and permeability prediction in tight sandstone reservoirs using optimized explainable ML frameworks \citep{liu2022permeability}. Comparative assessments have also shown ML models outperform or rival traditional physics-based approaches in water and groundwater modeling when sufficient data is available \citep{sheik2024machine}.

However, despite their high predictive performance, these models are often criticized for their “black-box” nature and the need for extensive, high-quality training data, which can hinder their interpretability and real-world application in sensitive water resource contexts \citep{eshetu2023interpretable, liu2022permeability}.

Despite advances in ML-based modeling, most studies continue to focus on prediction accuracy while ignnoring interpretability and process understanding \citep{10.3390/info13010015}. This gap limits the ability to uncover causal or mechanistic insights about the drivers of RWT across space and depth. 

To bridge this gap, this study applies Kolmogorov–Arnold Networks (KANs), a symbolic regression framework recently applied in environmental modeling for improved interpretability and generalization \citep{liu2024baseflow,granata2024advanced,xu2024kolmogorov}. KANs are designed to derive explicit symbolic equations from input data, providing transparency in model structure and enhancing transferability \citep{10.1002/msd2.70004}. Unlike SHAP-based explanations or physics-guided neural networks (PGNNs), which approximate interpretability post hoc, KANs could learn functional forms consistent with theoretical principles. To our knowledge, no prior study has applied KANs to lake or reservoir temperature prediction. 

\begin{figure*}
	\centering
		\includegraphics[width=\textwidth]{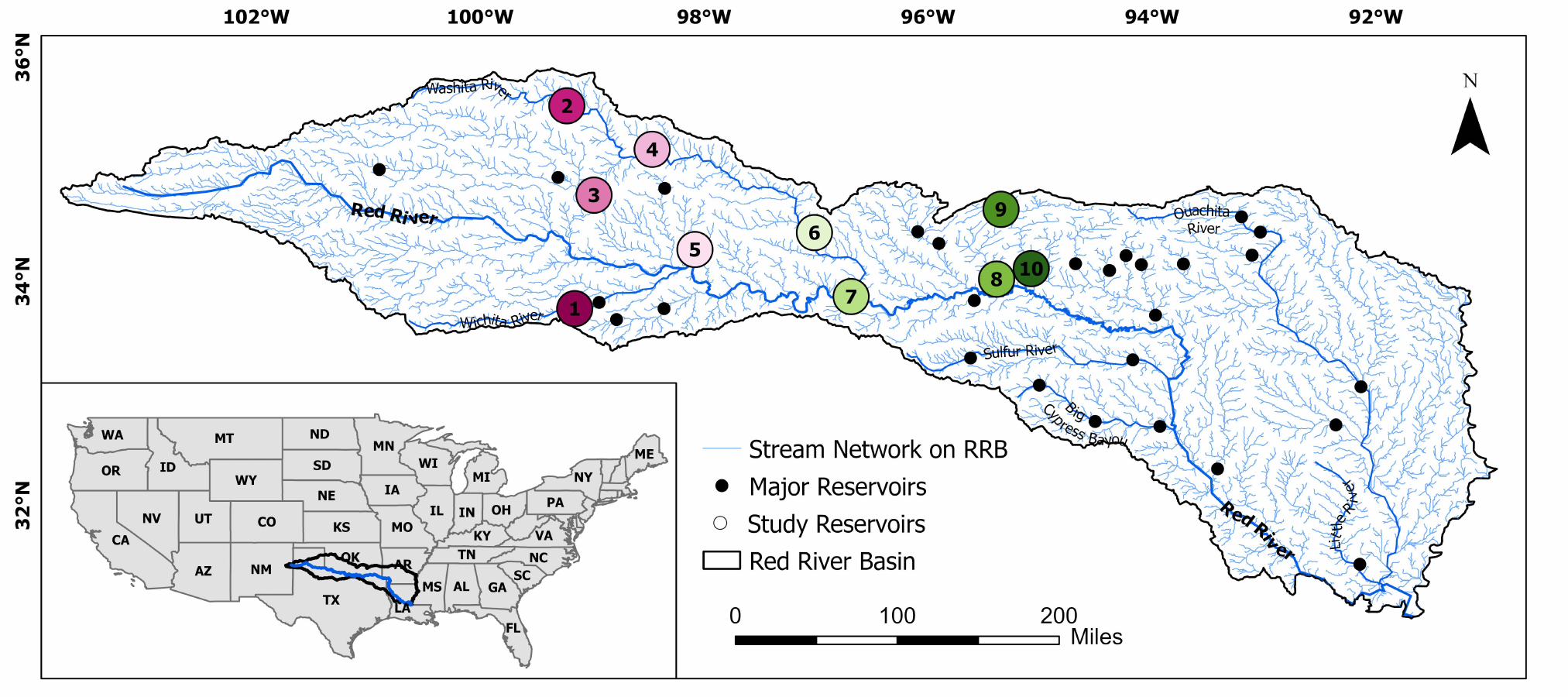}
	\caption{Map of study reservoirs within the Red River Basin divide. The main Red River channel is shown in dark blue, with major tributaries depicted in light blue. Lakes and reservoirs are represented by filled circles: black-filled circles indicate major reservoirs, while colored circles represent the major reservoirs that have available RWT measurements. The inset map highlights the basin's location in the south-central United States. A compact description of all studied reservoirs (labeled from 1-10 in this figure) is provided in Table \ref{reservoir_characteristics}.}
	\label{RRBmap}
\end{figure*}

This study focuses on 10 reservoirs distributed across the Red River Basin (RRB), a hydrologically complex watershed in the south-central United States, with a west-to-east gradient in water availability and water resources that support over three million people (Fig.\ref{RRBmap}; U.S. Census Bureau, 2023). The selected reservoirs represent a range of morphometric and climatic conditions across the basin. These were the only reservoirs for which consistent reservoir water temperature measurements could be obtained, albeit with variations in sampling frequency and depth profiles. Despite this limitation, the available data provide a valuable foundation to explore spatial and temporal patterns in RWT.

The overarching goal of this study is to develop a regionally transferable framework for predicting vertical temperature profiles and understanding their environmental drivers across diverse reservoirs. The study has three specific objectives:

1. Evaluate the performance of three ML models: RF, XGBoost, and MLP, in predicting RWT across different depths and temporal scales.

2. Identify the most influential environmental drivers of RWT using model-agnostic tools (e.g., SHAP) for the ML models developed.

3. Derive and interpret symbolic mathematical expressions from KAN to predict RWT, with increasing inputs based on feature importance identified in objective 2. This will help explain how environmental factors shape RWT variability, enhancing transparency, generalizability, and process insight across the basin.

Together, these objectives aim to advance both the predictive accuracy and interpretability of reservoir temperature modeling, offering a scalable toolset for water resource management in the face of climatic and hydrological uncertainty.

\section{Materials and Methods}

\subsection{Study Area}

The study reservoirs are located within the Red River Basin of the South in the United States, as illustrated in Fig.\ref{RRBmap}.  Within this region, there is a total of 38 major lakes and reservoirs \citep{ZAMANISABZI2019100638, roland2023}. In particular, this study focuses on 10 reservoirs that provide water supply, flood control, irrigation, and recreation. Table \ref{reservoir_characteristics} provides a synthesized description of these reservoirs, including their centroid coordinates, mean elevation, mean depths, mean volume, mean surface area, and main purpose. Reservoir mean depths range between 3.3 m and 10.9 m while their volumes at conservation pool range between \( 6.63 \times 10^7 \) and \( 2.89 \times 10^9 \) millions of m$^3$. Out of this list, Lake Texoma is one of the largest multipurpose reservoirs that supports hydroelectric power and navigation. Reservoirs in the Red River Basin are generally characterized by high turbidity due to suspended sediment inputs from tributaries, particularly during low flow conditions \citep{usgs1997redriver}. In addition, salinity is a major water quality concern in the region, largely driven by natural salt springs in the upper basin, which contribute elevated concentrations of dissolved solids, primarily chloride, sometimes exceeding seawater levels \citep{keller1988}.

The Red River originates in New Mexico and Texas, flows along the Texas-Oklahoma border, crosses Arkansas, and drains into the Mississippi River in Louisiana. It is the fourth-longest river in the United States, with a principal stream extending approximately 1,276 miles and a drainage area of approximately 233,000 km\textsuperscript{2} \citep{Christman2018}.

\begin{table}[htbp]
\small
\caption{Key characteristics of the 10 study reservoirs, listed from west to east. The data includes mean depth, surface area, volume, elevation (above mean sea level), and the primary purpose of each reservoir. }
\label{reservoir_characteristics}
\begin{tabularx}{\textwidth}{@{\extracolsep{\fill}} p{3.8cm} p{2.5cm} p{1.2cm} p{1.5cm} p{1.7cm} p{1.45cm} p{1.9cm}  @{}}
\toprule

Reservoir & Coordinates (centroid) & Mean Depth (m) & Surface Area* (km$^2$) & Volume* (m$^3$) &  Elevation* (AMSL)  &  Water-Use Sector (Main) \\
\midrule
(1) Lake Kemp & {\scriptsize (33.7393, -99.2319)} & 4.7 & 63.09 & $3.10 \times 10^8$ & 348.7 & Irrigation \\
(2) Foss Lake & {\scriptsize (35.5393, -99.1890)} & 7.0 & 24.35 & $2.09 \times 10^8$ & 500.6 & Municipal   \\
(3) Tom Steed Lake & {\scriptsize (34.7499, -98.9683)} & 3.9 & 25.90 & $1.21\times 10^8$ & 430.1 & Muncipal\\
(4) Ft. Cobb Reservoir& {\scriptsize (35.1621, -98.4568)} & 5.7 & 14.16 & $8.84\times 10^7$  & 409.1 & Municipal\\
(5) Waurika Lake & {\scriptsize (34.2348, -98.0469)} & 5.6 & 40.45 & $2.34\times 10^8$ & 289.9 & Municipal\\
(6) Arbuckle Lake & {\scriptsize (34.4331, -97.0266)} & 9.1 & 9.51  & $8.93\times 10^7$& 265.8 & Municipal\\
(7) Lake Texoma & {\scriptsize (33.8299, -96.5767)} & 10.9 & 300.2 & $2.89\times 10^9$ & 187.5 & Municipal\\
(8) Hugo Lake & {\scriptsize (34.0120, -95.3833)} & 3.3 & 46.9 & $1.74\times 10^8$ & 123.1  & Municipal\\
(9) Sardis Lake & {\scriptsize (34.6320, -95.3508)} & 5.2 & 55.07 & $3.32\times 10^8$  & 182.6 & Municipal\\
(10) Pine Creek Lake & {\scriptsize (34.1132, -95.0785)} & 3.0 & 15.16 & $6.63\times 10^7$ & 133.5 & Industrial\\
\bottomrule
\end{tabularx}
\vspace{-0.3cm} 
\begin{flushleft}
\textit{Note:} *Surface area, *elevation and *volume values are provided at conservation pool.
\end{flushleft}
\end{table}

The basin experiences diverse climatic conditions, with a marked gradient in water availability from west (driest) to east (wettest) \citep{gao2024}. In the upper basin, which spans both Texas and Oklahoma and is characterized by drier conditions, average annual precipitation ranges from approximately 870 mm to 920 mm. This region primarily utilizes reservoirs for water storage and to meet year-round water demands \citep{Wineland2022}. In contrast, the lower basin experiences a wetter climate, with average annual precipitation increasing to around 1,260 mm in the Arkansas portion and up to 1,590 mm in the Louisiana portion, with reservoirs primarily used for flood control \citep{roland2023}.

\subsection{Reservoir Water Temperature Dataset}

In situ measurements of RWT profiles were utilized to train and test machine learning models. These profiles were obtained through an online request to the National Water Quality Monitoring Council (NWQMC) via their website \citep{EPA_NWQMC_2025}. Profiles were measured at irregular time intervals spanning from 1996 to 2020 for the 10 reservoirs shown in Figure \ref{RRBmap} and Table \ref{reservoir_characteristics}. In instances where multiple discrete measurements were taken on a given lake-day (considering that a lake may have different measurement sites), these observations were treated independently. A total of 10,386 data points were collected, representing 674 RWT profiles across 10 reservoirs within the RRB. It is a prerequisite that each profile includes a minimum of four measurements to ensure data reliability.

Lake Texoma stands out with 490 measured profiles. Other reservoirs have moderate numbers, including Fort Cobb (40), Waurika Lake (33), Foss (27), Tom Steed and Pine Creek (19 each), Hugo (17), and Arbuckle (15) and Sardis (10). The most limited dataset is found in Sardis and Lake Kemp with just 2 profiles (see Figure \ref{RRBmap} and Table \ref{reservoir_characteristics}).

\begin{figure*}
	\centering
		\includegraphics[width=1\textwidth]{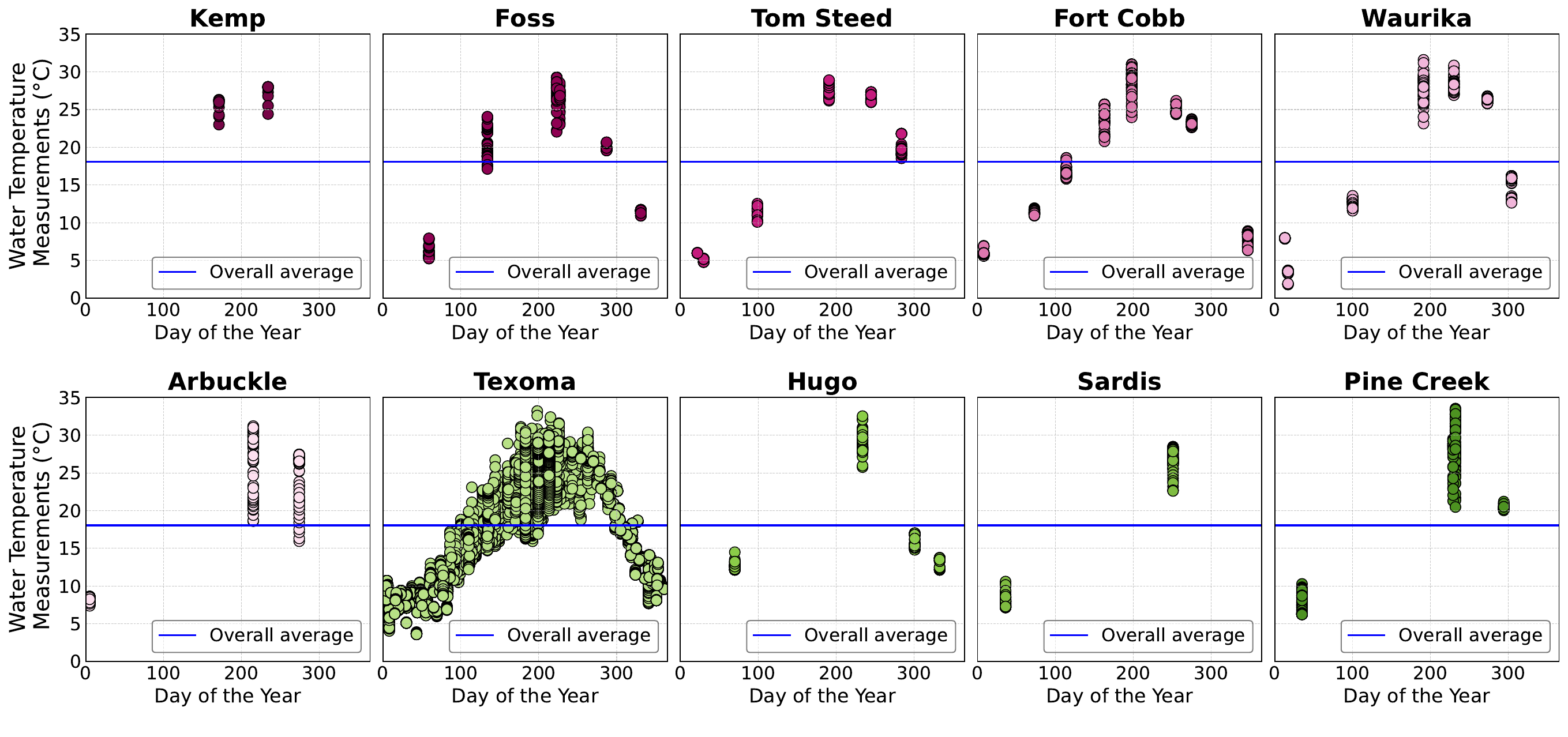}
	\caption{Intra-anual distribution of temperature measurements across reservoirs during the study period (1996-2020). Each point on the plot represents a single temperature measurement recorded on a specific day of the year and depth for a particular reservoir.  The blue dashed line represents the inter-annual daily average temperature for each reservoir, while the red solid line shows the overall inter-annual average temperature across all 10 reservoirs.}
	\label{dataDistr}
\end{figure*}

Figure \ref{dataDistr} illustrates the seasonal distribution of RWT measurements per reservoir. Among the 674 RWT profiles in the dataset, distinct seasonal distributions are observed with 173 profiles corresponding to the fall season (25.6\%), 140 profiles to the spring season (20.7\%), 181 profiles to the summer season (26.8\%), and 180 profiles to the winter season (26.7\%). Within this dataset, 218 profiles, constituting approximately 32\% of the total profiles, exhibit thermal stratification. A lake is considered stratified when the temperature difference between the epilimnion and the hypolimnion exceeds 1°C \citep{stefan1996simulated,foley2012long}. 

\vspace{0.3cm}

\subsection{Multiscale Environmental Predictors and Data Sources}
Accurate feature selection is crucial for developing robust and generalizable ML models. In this study, the choice of input variables was guided by two key principles: (1) relevance to the physical processes governing heat exchange, hydrodynamics, and stratification in lakes, and (2) data availability and quality across the study domain. This physics-informed selection approach ensures that the models are not only statistically sound but also grounded in domain knowledge of lake thermodynamics. Specifically, the covariate selection accounted for environmental, hydrologic, geomorphologic, and atmospheric factors known to influence reservoir water temperature dynamics. Additionally, the selection of variables was based on their availability in both observational datasets and future climate projections (CMIP6; 2020–2100) \citep{Fovargue_2021,ZAMANISABZI2019100638}. Table \ref{TableFeatures} provides a summary of the predictors used for the data-driven estimation of RWT. While some covariates were measured directly at the lake scale, others were derived from coarser 4-km resolution gridded products such as GridMET, which have been extensively validated \citep{abatzoglou2013}. We do not anticipate limitations arising from this multiscale integration; on the contrary, the low expected intra-pixel variability over lake surfaces and the broad availability of these predictors may enhance the model’s transferability across diverse lake systems.

\textbf{Air temperature (air\_temp; Table 2)} is a primary driver of RWT, as it governs the heat exchange between the atmosphere and the lake surface \citep{SCHMID2022467}. It is essential to note that lakes, especially large or deep ones, possess significant thermal inertia \citep{Piccolroaz}, meaning they do not respond instantaneously to changes in atmospheric conditions. Instead, their temperature responds with a delay that reflects the integration of heat over time. Consequently, the effects of air temperature are not limited to the current day, but also depend on conditions over preceding days. To capture this lagged response, we used a \textbf{seven-day (one-week; air\_temp7d; Table 2) moving average of air temperature} as a predictor. This temporal window was chosen as a balanced representation of short-term thermal memory: it is long enough to reflect the cumulative heating or cooling trends relevant to lake thermodynamics, but not so long as to dilute recent variations. Compared to shorter windows (e.g., 3 days), a one-week average better captures the smoothing effect of thermal inertia. Conversely, longer windows (e.g., 9 days) risk incorporating outdated influences, especially under rapidly changing atmospheric conditions. This choice is therefore both physically justified and empirically practical for modeling RWT dynamics.

When wind speed increases, water surface mixing in the epilimnion increases due to increased surface momentum \citep{Piccolroaz}. As a result, warmer surface water and cooler subsurface water can mix more efficiently. While wind promotes mixing in the epilimnion, its effect on deeper layers (hypolimnion) can be limited, especially in deep lakes. When the wind is calm (decrease in wind speed), it will result in less latent and sensible heat loss due to reduced evaporation and convection, and thus warming at the lake surface \citep{https://doi.org/10.1029/2019GL082752}. Under these circumstances, temperature stratification can occur, with warmer water at the surface and cooler water below. In such conditions, the air temperature appears to exert less influence on the lake's surface temperature. The reason to consider both the \textbf{1-day and 1-week (wind and wind\_avg7; Table 2)} average values is the cumulative effect of many days under consecutive wind-forced water mixing and their effect on RWT values' persistence (or change).

Direct precipitation onto the lake's surface affects RWT by introducing cooler water from rainfall, hail, or snow \citep{https://doi.org/10.1029/2019JC015950}. It also influences the lake's thermal structure, especially when near-frozen water or snow forms a surface layer of different temperatures. Since RWT can respond to precipitation in a delayed manner, 24-hour and 1-week average antecedent precipitation values \textbf{(prcp and prcp\_cum7; Table 2)} were added as driving input features.

Additionally, the amount of water stored in each lake \textbf{(vol\_lake; Table 2)} influences the thermal inertia, currents, and dynamic structure of the lake \citep{hess-21-6253-2017}. Larger lakes with higher volumes tend to change temperature more slowly than smaller ones. A substantial water inflow \textbf{(via streamflow; inflow\_lake; Table 2)}, coupled with a short residence time (which refers to the average amount of time water remains within a lake before flowing out through its outlet), means that new water mixes relatively quickly with the lake's existing water, leading to rapid adjustments in temperature. This can result in more significant temperature fluctuations, making the lake highly responsive to external temperature changes \citep{christianson2020compound,hipsey2014glm}. Conversely, limited water inflow, combined with a longer residence time, creates a more stable thermal environment, as the lake retains its heat for an extended period and is less sensitive to short-term temperature variations. Thus, the interplay between water inflow and residence time governs a lake's ecosystem's thermal resilience and responsiveness to changing environmental conditions \citep{christianson2020compound, hipsey2014glm}. 

Depth \textbf{(depth\_measure; Table 2)} is another key variable influencing lake thermal dynamics. It affects air–water interactions, such as how solar radiation, wind energy, and precipitation are absorbed and distributed. Greater depth is associated with enhanced stratification stability, distinct thermocline formation, and more layered water currents, which collectively shape thermal profiles and biological processes within the lake.

The surface area-to-maximum depth ratio \textbf{(surf\_area\_depth; Table 2)} provides important information about lake morphometry, influencing the extent and efficiency of heat exchange with the surrounding environment \citep{hess-21-6253-2017}. Shallow, wide lakes may warm or cool more rapidly than narrow, deep ones due to this geometric relationship. In addition to these dynamic features, static or contextual attributes, such as mean depth, are included as time-invariant features in the model to capture baseline differences across reservoirs.

\begin{table}
\small
\caption{Input features used by the ML models to predict reservoir water temperature. Note that the input features drive the lake temperature in some way by providing encoded information about the relations between inputs and outputs.}
\label{TableFeatures}
\begin{tabularx}{\textwidth}{p{5.1cm}  >{\footnotesize}p{2.4cm} >{\footnotesize}p{3.9cm}  >{\footnotesize}p{4.2cm}
}
\toprule
Feature & Abbrev. & Source and Resolution & Range (Min, Max) \\
\midrule
Same day air temperature (AT)& air\_temp & GridMET (4 km, daily) & ($-6.91$, 34.35) °C \\
7-day antecedent average AT & air\_temp7d & GridMET (4 km, daily) & ($-2.81$, 34.54) °C \\
Same day precipitation (P) & prcp & GridMET (4 km, daily) & (0, 67.20) mm \\
7-day antecedent cumulative P & prcp\_cum7 & GridMET (4 km, daily) & (0, 33.27) mm \\
Same day wind speed (WS)  & wind & GridMET (4 km, daily) & (1.30, 8.72) m/s \\
7-day antecedent average WS & wind\_avg7 & GridMET (4 km, daily) & (2.41, 6.61) m/s \\
Lake volume & vol\_lake & USACE (daily) & (31,956, \small $5.85 \times 10^{6}$)AF \\
Water inflow & inflow\_lake & USACE (daily) & (0, 131,300) 24-hr Avg CFS\\
Measurement depth & depth\_measure & NWQMC (Point, instant) & (0, 30.48) m \\
Surface area max depth ratio & surf\_area\_depth & Calculated & (235.67, 14,775) $\mathrm{m^2\!/\!m}$
 \\
\bottomrule
\end{tabularx}
\end{table}

\vspace{0.4cm}

\subsection{Machine Learning Models}

This study employs a suite of ML models, ranging from classical algorithms to deep learning architectures and emerging symbolic frameworks such as Kolmogorov–Arnold Networks, to predict RWT across the Red River Basin.

\subsubsection{Random Forest}
Random Forest is an ensemble model that combines multiple decision trees to improve prediction accuracy and reduce overfitting \citep{breiman2001random}. It constructs a forest of decision trees, where each tree is trained on a random subset of input variables and samples (bagging). RF is well-suited for capturing complex and nonlinear relationships between input and output variables and has been successfully applied to reservoir water temperature prediction \citep{w11050910}. 

The general RF prediction is given by Equation (\ref{RF_eq}), where $f(x)$ is the ensemble output, $T(x; \theta_b)$ is the prediction of the $b^\text{th}$ tree, $B$ is the total number of trees, and $\theta_b$ represents the model parameters (e.g., node split points and thresholds) learned during training for tree $b$.

\begin{equation}\label{RF_eq}
f(x) = \frac{1}{B} \sum_{b=1}^B T(x;\theta_b)
\end{equation}

In addition to the model parameters learned during training, RF has several hyperparameters that are set prior to training and govern the learning process. Tuning these hyperparameters can significantly affect model performance. Key hyperparameters include:

(1) the maximum depth of each tree,
(2) the minimum number of samples required at a leaf node,
(3) the maximum number of features considered for splitting at each node, and
(4) the total number of trees in the forest \citep{breiman2001random}.

\vspace{0.2cm}

\subsubsection{Gradient Boosting}
Gradient boosting is an ensemble learning technique that builds predictive models by sequentially adding weak learners, typically shallow decision trees, to correct the errors made by previous models \citep{friedman2001greedy}. Unlike bagging methods such as Random Forest, which train trees independently in parallel and aggregate their predictions, boosting methods construct trees sequentially, with each new tree trained to minimize the residual error of the ensemble up to that point. The idea is to iteratively refine the prediction by fitting new models to the gradients (i.e., the negative gradients of the loss function), hence the name gradient boosting.

One of the most efficient and widely used implementations of gradient boosting is Extreme Gradient Boosting (XGBoost) \citep{XGBoost}. XGBoost builds upon the foundational principles of gradient boosting and introduces several enhancements to improve efficiency, scalability, and generalization performance. These enhancements include parallelized tree construction, regularization to reduce overfitting, and optimized handling of sparse data and missing values.

In XGBoost models, the target variable y$_i$ is modeled as the sum of predictions from an ensemble of regression trees f$_j$ as shown in equation (\ref{XGBoost_eq}).

\begin{equation}
    y_i = \sum_{j=1}^m f_j(x_i) + \epsilon_i\label{XGBoost_eq}
\end{equation}

Where $x_i$ is the input vector for the $i^{\text{th}}$ observation, $m$ is the number of trees (or boosting rounds), and $\epsilon_i$ is the residual error not captured by the model.

Each tree f$_j$ is trained to minimize a regularized objective function that balances model fit and complexity. Unlike traditional gradient boosting, XGBoost explicitly incorporates L1 and L2 regularization into the objective function, helping to prevent overfitting, especially in high-dimensional settings.

Similar to other machine learning models, XGBoost has several hyperparameters that can be tuned to optimize performance. Key hyperparameters include:

(1) Gamma: the minimum loss reduction required to make a further partition on a leaf node; (2) Subsample ratio of columns by tree: controls feature sampling to reduce overfitting; (3) Maximum tree depth: limits model complexity; (4) Minimum child weight: the minimum sum of instance weights in a child node; (5) Learning rate (eta): controls the contribution of each tree; (6) Number of estimators: the total number of boosting rounds (i.e., trees in the ensemble).

These hyperparameters are not learned during training but must be specified or searched via an optimization mechanism before model fitting. They play a critical role in balancing model bias and variance.

\subsubsection{Multilayer Perceptron}
The multi-layer perceptron is a type of artificial neural network (ANN) composed of multiple layers of interconnected nodes (neurons) that transform input data through a series of weighted linear combinations and nonlinear activation functions \citep{rumelhart1986learning}. MLPs are capable of learning complex relationships between input and output variables and are well-suited for modeling nonlinear and high-dimensional data. They have been successfully applied to a range of prediction tasks, including RWT forecasting \citep{MAIER2000101}. The basic structure of an MLP is described in Equation (\ref{ANN_eq}), where y is the predicted output, f is the activation function, w$_i$ are the weights, x$_i$ are the input features, b is the bias term, and n is the number of inputs. The weights and biases are optimized during training to minimize prediction error and improve the model’s performance.

\begin{equation}\label{ANN_eq}
y = f\left(\sum_{i=1}^n w_i x_i + b\right)
\end{equation}

An MLP model consists of parameters, such as the weights and biases, that are learned during training through optimization algorithms like stochastic gradient descent. In addition to these trainable parameters, the model’s performance is heavily influenced by several hyperparameters, which must be specified prior to training and tuned to achieve optimal results. Key hyperparameters include: (1) the number of hidden layers, (2) the number of neurons per layer, (3) the batch size, (4) the number of training epochs, (5) the dropout rate used for regularization, and (6) the learning rate that controls the speed of weight updates. These hyperparameters govern the architecture and learning behavior of the model. When the network consists of three or more layers (including input and output layers), it is commonly referred to as a deep neural network or a deep learning (DL) model.

\vspace{0.2cm}
\subsection{Training, Testing and Model Evaluation}

The dataset was randomly divided into a training set (70\%) and a testing set (30\%) at the temperature profile level, where each profile represents a complete vertical temperature measurement at a specific time and location. A fixed random seed (random\_state=42) was used to ensure reproducibility. All observations within each profile (i.e., measurements across depths) were kept together during the split to prevent data leakage. Within the training set, 5-fold cross-validation was applied to optimize model performance and reduce overfitting. This strategy allowed for robust hyperparameter tuning while preserving a large, independent testing set for final model evaluation. For reservoirs with only two available temperature profiles, one was assigned to the training set and the other to the testing set, ensuring that even sparsely sampled lakes contributed to both model development and validation. However, the trustworthiness of model performance estimates in such cases may be limited, as the small number of profiles reduces the statistical robustness of both training and testing. These results should therefore be interpreted with caution, especially when generalizing to similarly undersampled systems.

Given the moderate size of the dataset (\textasciitilde10,000 instances and 10 covariates), a randomized search approach was chosen for hyperparameter optimization as it provides an efficient balance between exploration of the parameter space and computational cost. An initial sensitivity analysis helped constrain the search to meaningful parameter ranges, reducing the risk of overfitting while ensuring that each model configuration was adequately tuned for predictive performance. All input features were properly standardized using min-max scaling, which rescales each feature to a fixed range [0,1], preserving the distribution shape while ensuring comparability across variables. 

Model accuracy was evaluated using Root Mean Squared Error (RMSE), Mean Absolute Error (MAE), and the coefficient of determination ($R^2$). RMSE quantifies the average magnitude of the prediction error by penalizing larger errors more heavily, while MAE measures the average absolute difference between predicted and actual values; both are ideally close to 0. The $R^2$ score reflects the proportion of variance in the dependent variable explained by the model, with values closer to 1 indicating a better fit.

\vspace{0.5cm}
\subsection{SHAP-Based Feature Importance Analysis}

To interpret the contribution of each predictor to the RWT model (RF, XGBoost, MLP) output, we used SHAP (SHapley Additive exPlanations), a model-agnostic approach grounded in cooperative game theory. Two types of SHAP visualizations were generated to facilitate interpretation: (1) summary plots, which rank predictors by their overall importance and illustrate the direction and magnitude of their effects; and (2) heatmap plots, which show the relationship between individual predictor values and their corresponding SHAP values, highlighting interaction effects.

SHAP is used to explain the output of complex machine learning models. It connects optimal credit allocation with local explanations using classic Shapley values from game theory and their related extensions \citep{NIPS2017_7062}. The computation of SHAP values involves evaluating the model's output for all possible combinations of feature values and averaging the contributions of each feature across these combinations. This process ensures that each feature is assigned a fair share of the credit for the model's prediction \citep{NIPS2017_7062}. Given input variables and a trained model \( F \), SHAP can utilize a model \( G \) to assess the contribution of each input variable to model \( F \). The model \( G \) is defined as follows:

\begin{equation}
G(z) = \phi_0 + \sum_{i=1}^{q} \phi_i t_i,
\end{equation}

where \( q \) is the number of input variables, \( t_i \) is the simplification of the input variable (the \( t_i \) value corresponding to the feature used for data prediction is 1, and the \( t \) value corresponding to the unused feature is 0), and \( \phi_i \in \mathbb{R} \) represents the contribution of each feature to the model. The contribution function \( \phi \) is as follows (Equation \ref{eqi}):

\begin{equation}
\label{eqi}
\phi_i(F, x) = \sum_{t \subseteq x} \frac{|t|!(q - |t| - 1)!}{q!} \left[ F(t) - F(t \setminus i) \right],
\end{equation}
\vspace{0.3cm}

SHAP can show the impact of each sample feature, highlighting whether that influence is positive or negative for the model \citep{WANG2022127320}.

To assess and compare the influence of input variables across machine learning models, we used SHAP applied to the RF, XGBoost, MLP models to extract a consistent, interpretable ranking of feature importance. Overall, SHAP enabled both robust model diagnostics and a principled method for reducing model complexity, facilitating both feature selection and equation discovery across hybrid ML-symbolic workflows.

SHAP was chosen over permutation importance and other feature ranking methods because it provides a consistent, model-agnostic framework with both global and local interpretability. Compared to permutation importance, SHAP offers fine-grained input attribution and satisfies key theoretical properties such as consistency and local accuracy \citep{10.1007/s10822-020-00314-0}. 

\vspace{0.2cm}
\subsection{Kolmogorov-Arnold Networks (KANs) for Symbolic Equation Discovery}

After evaluating the predictive performance of the RF, MLP, and XGBoost models, SHAP analysis was conducted on all three to identify the most informative predictors. However, only the SHAP-derived feature importance ranking from the top-performing model was used to guide the incremental input selection for the KAN models. KAN was then applied to derive symbolic equations for predicting RWT. Two sets of ten KAN models were developed: Set 1, consisting of ten \textit{simple} (interpretable) models, and Set 2, consisting of ten \textit{complex} models with higher accuracy but reduced interpretability. The primary objective was to assess the trade-off between model complexity (in terms of the number of input variables) and predictive accuracy by gradually increasing the number of inputs used, following the SHAP-ranked feature importance. 

Here, it is critical to distinguish between two forms of complexity: the number of input variables and the functional complexity of the resulting equation. The former refers to the number of features (e.g., temperature, depth, precipitation) included in the model, while the latter reflects the mathematical form of the equation, ranging from simple additive terms to nested nonlinear transformations. For example, a model with five inputs could yield either a straightforward linear equation or a highly nonlinear expression involving trigonometric and exponential terms, depending on the network architecture.

KANs are a type of neural network inspired by the Kolmogorov-Arnold representation theorem \citep{kolmogorov1957representation} designed to overcome the limitations of traditional multi-layer perceptrons by enabling the fitting of complex nonlinear relationships with a single layer \citep{Kan2025}. The theorem asserts that high-dimensional functions can be represented as compositions of one-dimensional functions. This feature allows KANs to be highly efficient in representing complex functions with fewer parameters.

According to the theorem, if \( f \) is a multivariate continuous function defined on a bounded domain, then \( f \) can be expressed as a finite composition of continuous functions of a single variable and the binary operation of addition. Specifically, for a smooth function \( f: [0, 1]^n \rightarrow \mathbb{R} \), it can be written as:

\begin{equation}
f(x_1) = f(x_1, \ldots, x_n) = \sum_{q=1}^{2n+1} \Phi_q \left( \sum_{p=1}^{n} \varphi_{q,p}(x_p) \right)
\end{equation}

where \( \varphi_{q,p}: [0, 1] \to \mathbb{R} \) and \( \Phi_q: \mathbb{R} \to \mathbb{R} \) are continuous functions of a single variable.
\vspace{0.3cm}

In terms of interpretability, KANs offer distinct advantages. Unlike traditional black-box models such as MLPs, which rely on fixed activation functions and linear weights, KANs replace these weights with learnable univariate functions applied to edges. As a result, KANs inherently promote interpretability by decomposing high-dimensional relationships into more transparent 1D functional components \citep{Kan2025}. 

This architecture enables clearer insight into how each edge contributes to the model’s behavior, facilitating direct human interpretability. While the univariate functions are typically approximated using flexible basis functions (e.g., splines), the interpretability stems from the univariate decomposition itself, rather than the specific approximation method. Moreover, KANs can be converted into symbolic expressions, offering explicit mathematical formulas that describe the relationships uncovered by the model. 

For the training and testing of the KAN models, the same number of data points (10,386) was utilized as in the RF, XGBoost, and MLP models. All input variables and the target variable were normalized using Min-Max scaling to the range [0, 1]. This normalization was applied to align the data with the activation function's output range, thereby improving learning efficiency and convergence stability. As a result of this transformation, the derived equations operate within a normalized feature space. The set of \textit{simple} models was calibrated using architectures with one layer and two hidden neurons, in contrast to the \textit{complex} versions, which had one to two layers and up to three hidden neurons. Both sets of model types were trained on the same inputs; however, the complex architectures were allowed to use more expressive function bases, including trigonometric, exponential, and rational components. This architectural flexibility enabled the discovery of highly nonlinear and compact symbolic forms, including nested expressions such as \(\cos(ax + b)\), \(\exp(cx)\), and other composite terms.

\begin{figure}
\begin{center}
\resizebox{1\textwidth}{!}
  {\includegraphics{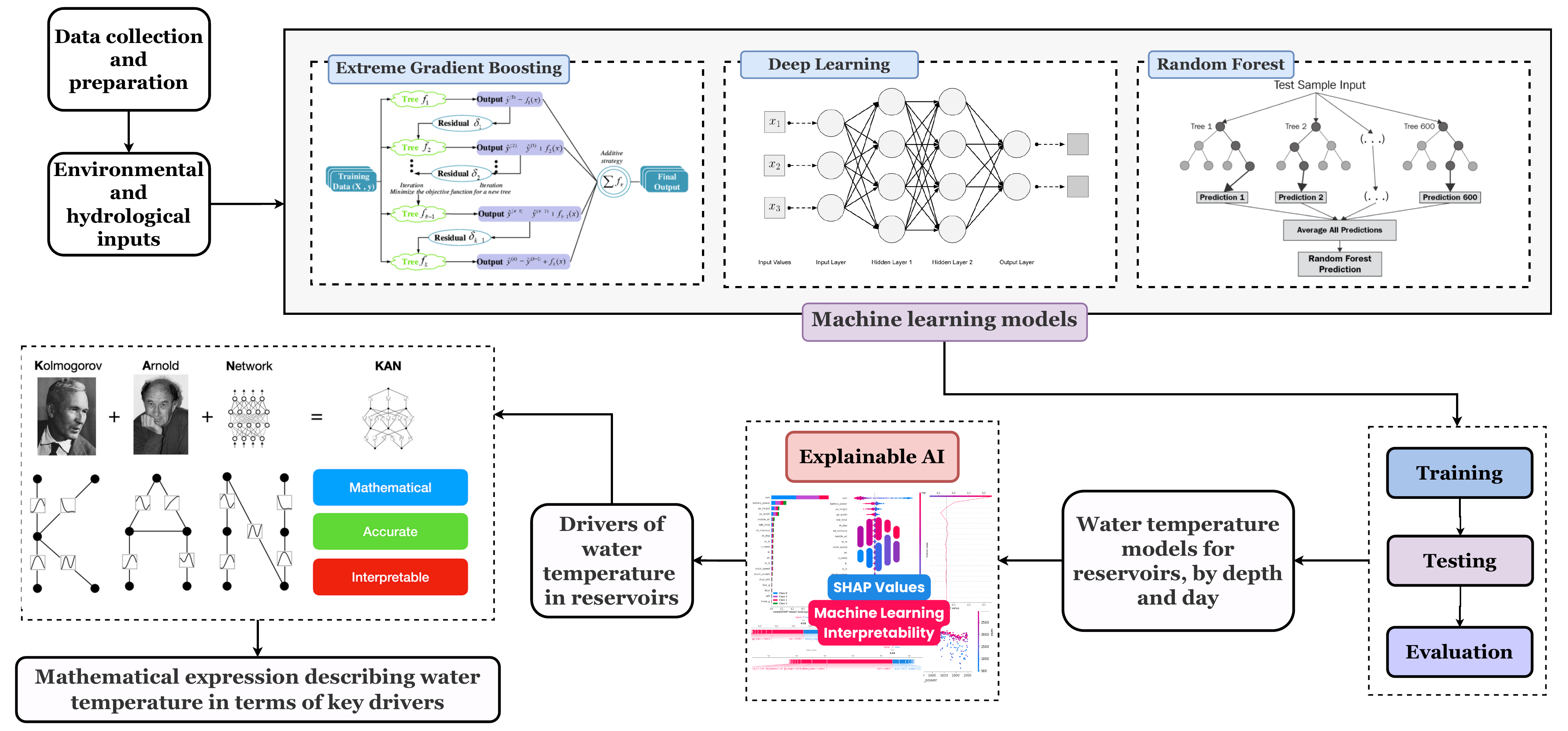}}
\end{center}
\caption{Summary of methodological framework from multi-source inputs to machine learning model training, testing and evaluation, SHAP feature ranking and Kolmogorov-Arnold Networks to distill predictive equations.}
\label{Methodology1}
\end{figure}

The schematic methodology framework shown in Figure \ref{Methodology1} provides a visual overview of the methodological steps described above.

\section{Results}

\subsection{Model Tuning, Performance Evaluation, and Comparative Analysis}

The RF, XGBoost and MLP hyperparameter optimization process resulted in distinct configurations for each model. The RF model achieved its best performance using 100 estimators, a maximum of 4 randomly sampled features, and a maximum tree' depth of 30. XGBoost performed optimally with 600 estimators, a learning rate of 0.01, a maximum depth of 9, a gamma of 0.3, and a subsample ratio of 1.0. For the MLP model, the best configuration included two hidden layers with 48 neurons each, a batch size of 32, 1000 training epochs, a dropout rate of 0.1, and a learning rate of 0.01. These configurations were then used to train the models for subsequent performance evaluation.

\begin{figure*}
	\centering
		\includegraphics[width=\textwidth]{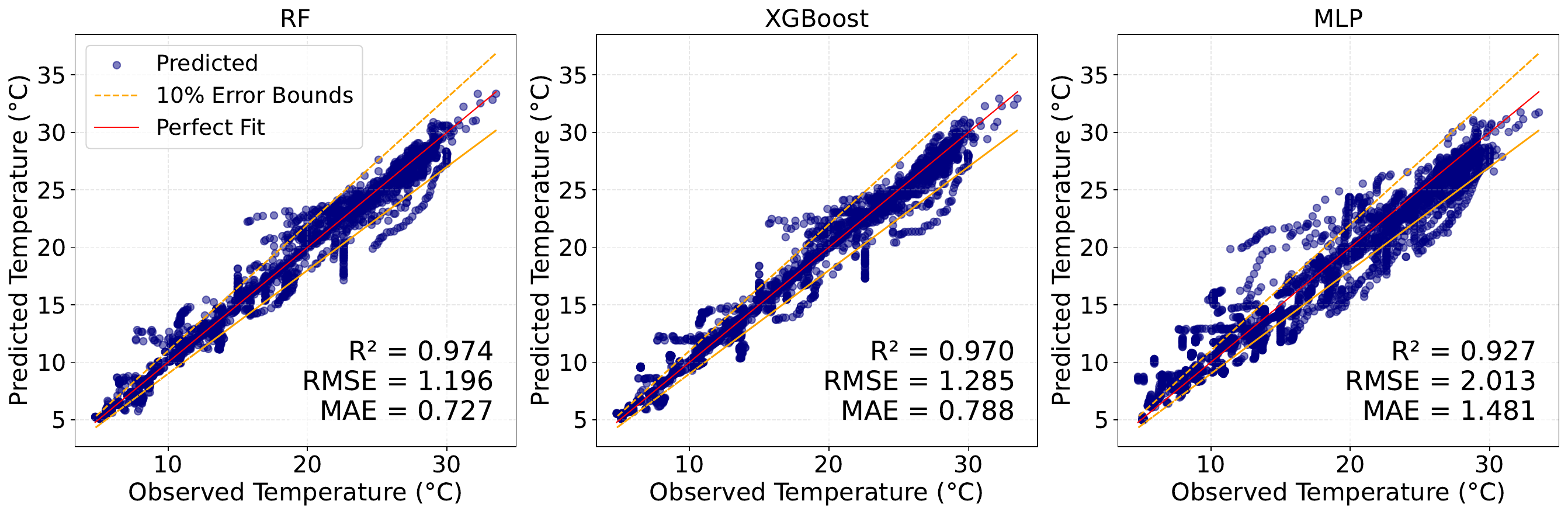}
	\caption{Scatterplots comparing observed and predicted RWT values for the (left) RF, (center) XGBoost, and (right) MLP models on the testing dataset. The 1:1 line indicates perfect agreement, while the dashed lines represent ±10\% error bounds. All models show strong alignment with the 1:1 line, indicating high predictive accuracy.}
	\label{combined_scatter_plots}
\end{figure*}

Figure \ref{combined_scatter_plots} presents scatterplots comparing predicted and observed reservoir water temperatures for the RF, XGBoost, and MLP models on the test set. In each plot, predicted values are plotted on the y-axis and observed values on the x-axis, with the 1:1 line indicating perfect agreement. All three models demonstrate strong alignment with this line, and most data points fall within the ±10\% error bounds, indicating that the models reliably capture the variability in reservoir water temperature across the testing dataset. Key performance metrics (RMSE, MAE, and $R^2$) are also reported in the bottom-right corner of each plot. Among the three, RF achieved the highest accuracy (RMSE = 1.196°C, MAE = 0.727°C, $R^2$ = 0.974), followed by XGBoost (RMSE = 1.285°C, MAE = 0.787°C, $R^2$ = 0.970) and MLP (RMSE = 1.830°C, MAE = 0.930°C, $R^2$ = 0.945). While all models demonstrated strong predictive skill, RF consistently outperformed the others, exhibiting lower errors and higher explained variance across all metrics.

An examination of model performance across the individual test reservoirs (Table~\ref{ResultsTEST_ap}) reveals both overall robustness and notable variability linked to site-specific conditions. Generally, all three machine learning approaches achieved high predictive skill across the studied reservoirs, with $R^2$ frequently exceeding 0.90 and RMSE values typically below 1.5$^\circ$C. The RF model consistently delivered exceptional results, achieving the highest $R^2$ values and lowest RMSE in most reservoirs, including Fort Cobb, Foss, Hugo, Texoma, Pine Creek, and Tom Steed. XGBoost exhibited competitive performance, occasionally matching or slightly surpassing RF on individual metrics, as seen in Arbuckle (highest $R^2$ of 0.991 and lowest RMSE of 0.703$^\circ$C), Sardis Lake (highest $R^2$ of 0.844 and lowest RMSE of 0.220$^\circ$C), and Waurika Lake (highest $R^2$ of 0.992 and lowest RMSE of 0.664$^\circ$C).

\vspace{0.5cm}

\begin{table*}
\small
\caption{Test model performance results by reservoir. For each reservoir and metric, the best-performing value among RF, XGBoost, and MLP is highlighted in \textbf{bold}. Specifically, the highest $R^2$ and the lowest RMSE values per reservoir are bolded to indicate superior predictive performance.}\label{ResultsTEST_ap}
\begin{tabular*}{\linewidth}{@{\extracolsep{\fill}} l c c c c c c c @{}}
\toprule
\textbf{Reservoir} & \multicolumn{2}{c}{\textbf{RF}} & \multicolumn{2}{c}{\textbf{XGBoost}} & \multicolumn{2}{c}{\textbf{MLP}} \\
 & $R^2$ & RMSE (°C) & $R^2$ & RMSE (°C) & $R^2$ & RMSE (°C) \\
\midrule
Arbuckle Reservoir   & 0.989 & 0.787 & \textbf{0.991} & \textbf{0.703} & 0.978 & 1.112 \\
Fort Cobb Reservoir  & \textbf{0.994} & \textbf{0.645} & 0.993 & 0.647 & 0.992 & 0.756 \\
Foss Reservoir       & \textbf{0.985} & \textbf{0.944} & 0.980 & 1.071 & 0.947 & 1.751 \\
Hugo Lake            & \textbf{0.598} & \textbf{0.987} & 0.572 & 1.019 & 0.524 & 1.074 \\
Lake Texoma          & \textbf{0.968 }& \textbf{1.283} & 0.963 & 1.381 & 0.920 & 2.034 \\
Pine Creek Lake      & \textbf{0.998} & \textbf{0.467} & 0.997 & 0.559 & 0.995 & 0.652 \\
Sardis Lake          & 0.363 & 0.445 & \textbf{0.844} & \textbf{0.220} & 0.761 & 0.273 \\
Tom Steed Reservoir  & \textbf{0.997} & \textbf{0.576} & 0.996 & 0.624 & 0.931 & 2.855 \\
Waurika Lake         & 0.989 & 0.764 & \textbf{0.992} & \textbf{0.664} & 0.855 & 2.760 \\
\bottomrule
\end{tabular*}
\end{table*}

\vspace{0.2cm}

Model performance, however, varied notably across sites. Hugo Lake and Sardis Lake emerged as outliers. While Hugo showed moderate performance across all three models, Sardis presented particularly poor results for RF ($R^2 = 0.363$). Interestingly, XGBoost achieved strong accuracy at Sardis ($R^2 = 0.844$, RMSE = 0.220$^\circ$C), while MLP also performed reasonably well ($R^2 = 0.761$, RMSE = 0.273$^\circ$C). Notably, Sardis had a severely limited dataset, consisting of only two temperature profiles: one used for training and one for testing.

In contrast, reservoirs such as Fort Cobb, Foss, and Tom Steed demonstrated uniformly strong agreement across all models, indicating more stable thermal regimes or more predictable responses to the selected drivers. Overall, tree-based ensemble methods, particularly RF, offered more consistent and accurate performance across a range of reservoir conditions, whereas MLP showed promise but also highlighted the importance of data quality and site-specific characteristics in determining model success.

\begin{figure}
	\centering
		\includegraphics[width=\textwidth]{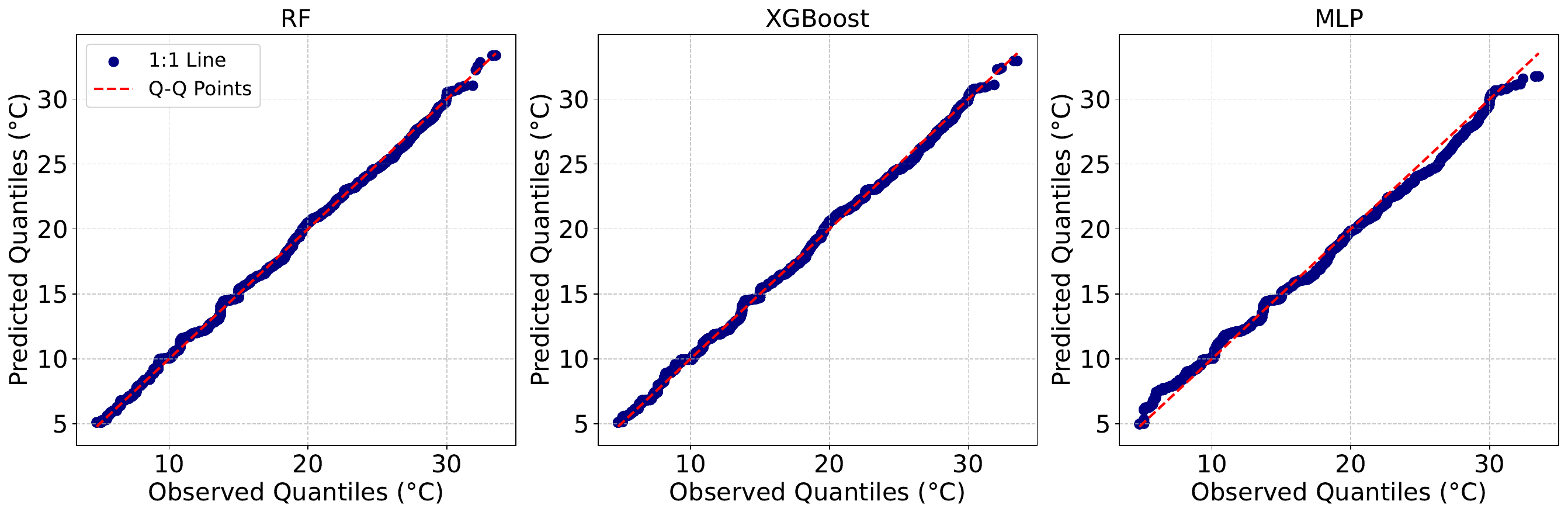}
	\caption{Quantile–Quantile (Q-Q) plots of observed vs. predicted values for the RF (left), XGBoost (center), and MLP (right) models on the test set. The red dashed line denotes the 1:1 reference line.}
	\label{combined_quantile_plots}
\end{figure}

The Q–Q plots in Figure \ref{combined_quantile_plots} compare the observed and predicted quantiles for the three models: RF, XGBoost, and MLP. All models demonstrate strong alignment with the 1:1 reference line (red dashed line), indicating that the predicted distributions closely match the observed temperature values. In the lower temperature range, RF and XGBoost show slightly better agreement than MLP, with their quantiles more tightly clustered around the reference line. The MLP model exhibits a few minor deviations in this region, with some points deviating slightly from the ideal line. At the higher end of the temperature distribution, all models show modest discrepancies: the MLP model tends to slightly underestimate extreme values, while RF and XGBoost display small under- or overestimations at a few points. Overall, the deviations observed are limited, and all three models effectively capture the overall distribution of the target variable, with RF and XGBoost exhibiting marginally better performance, particularly in the tails.

\vspace{0.4cm}

\subsection{Interpreting Global Feature Contributions to Reservoir Water Temperature Predictions}

\begin{figure}
  \centering
  \subfloat[RF]{\includegraphics[width=0.75\linewidth]{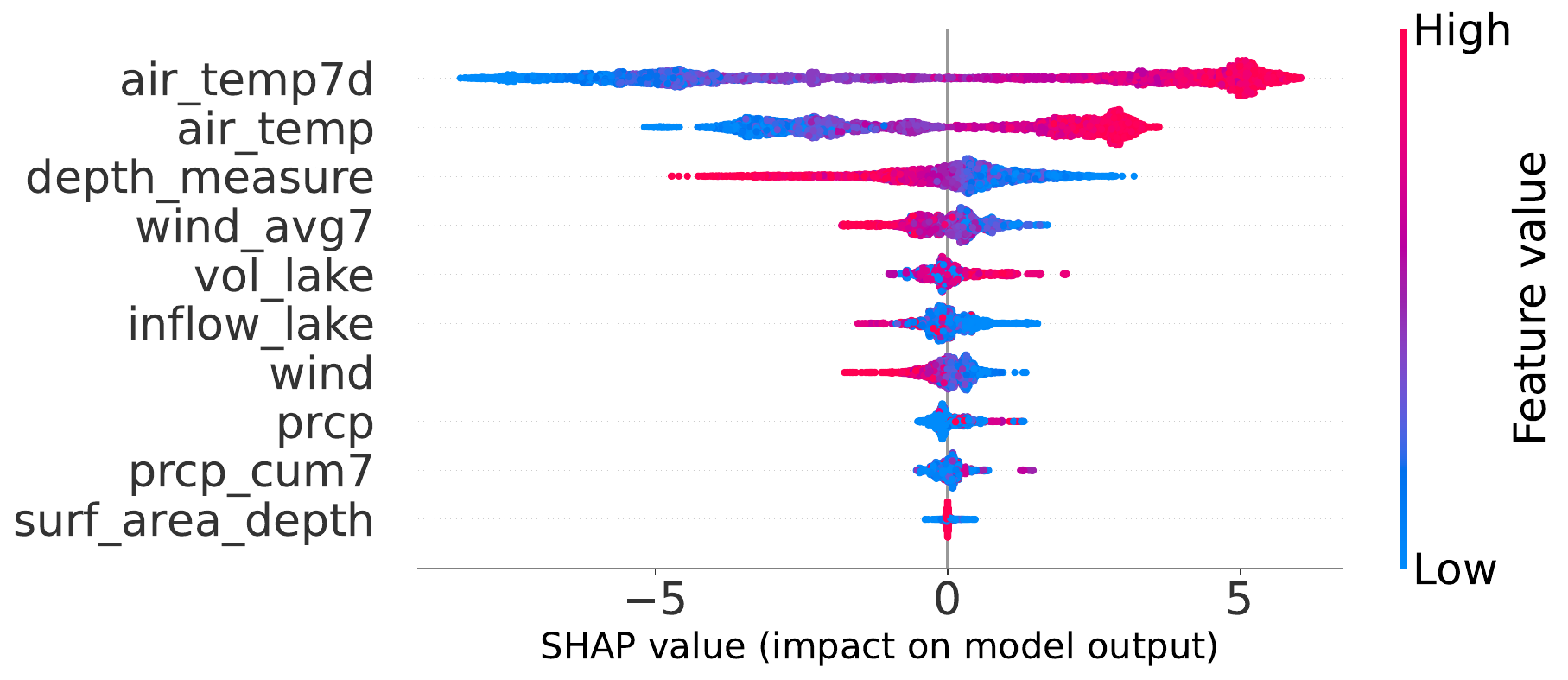}}\hfill

  \vspace{0.5em}  

  \subfloat[XGBoost]{\includegraphics[width=0.75\linewidth]{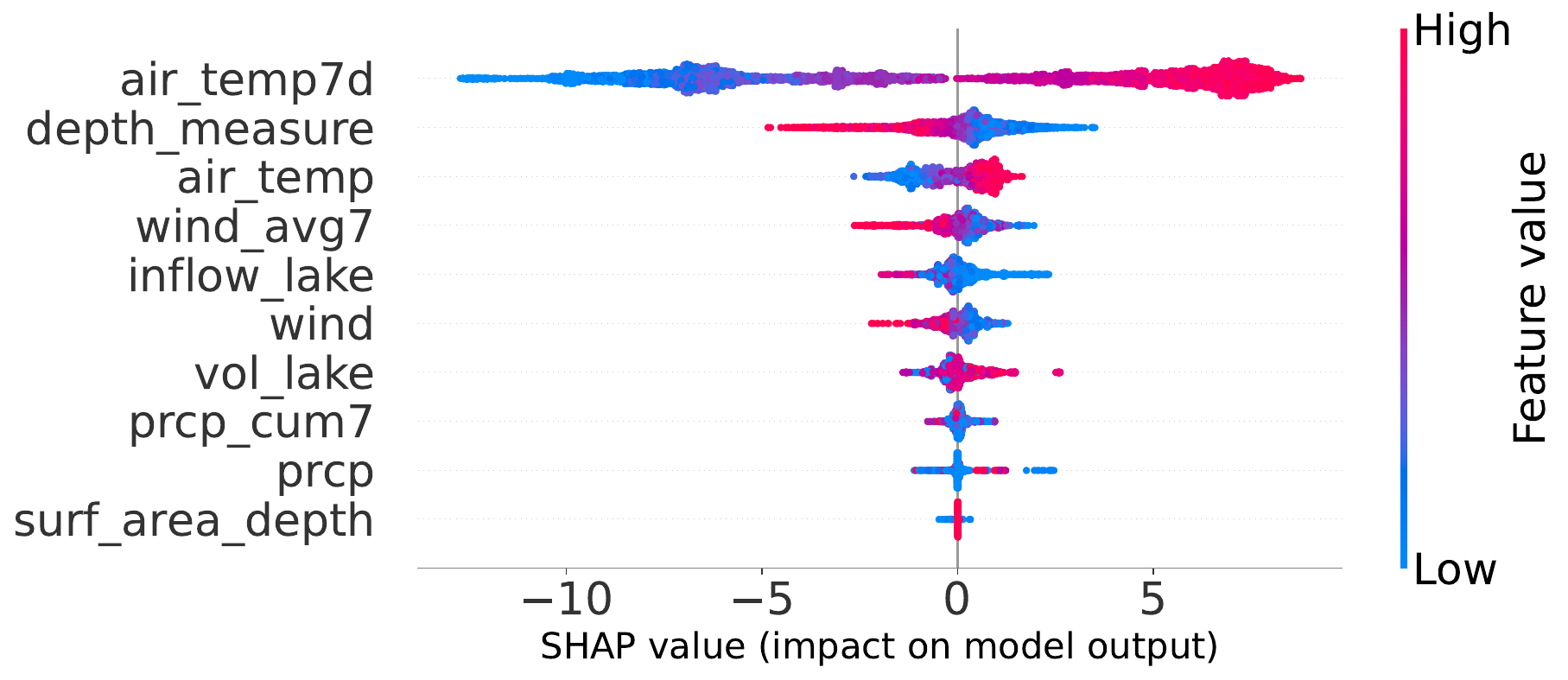}}\hfill

  \vspace{0.5em}
  \subfloat[MLP]{\includegraphics[width=0.75\linewidth]{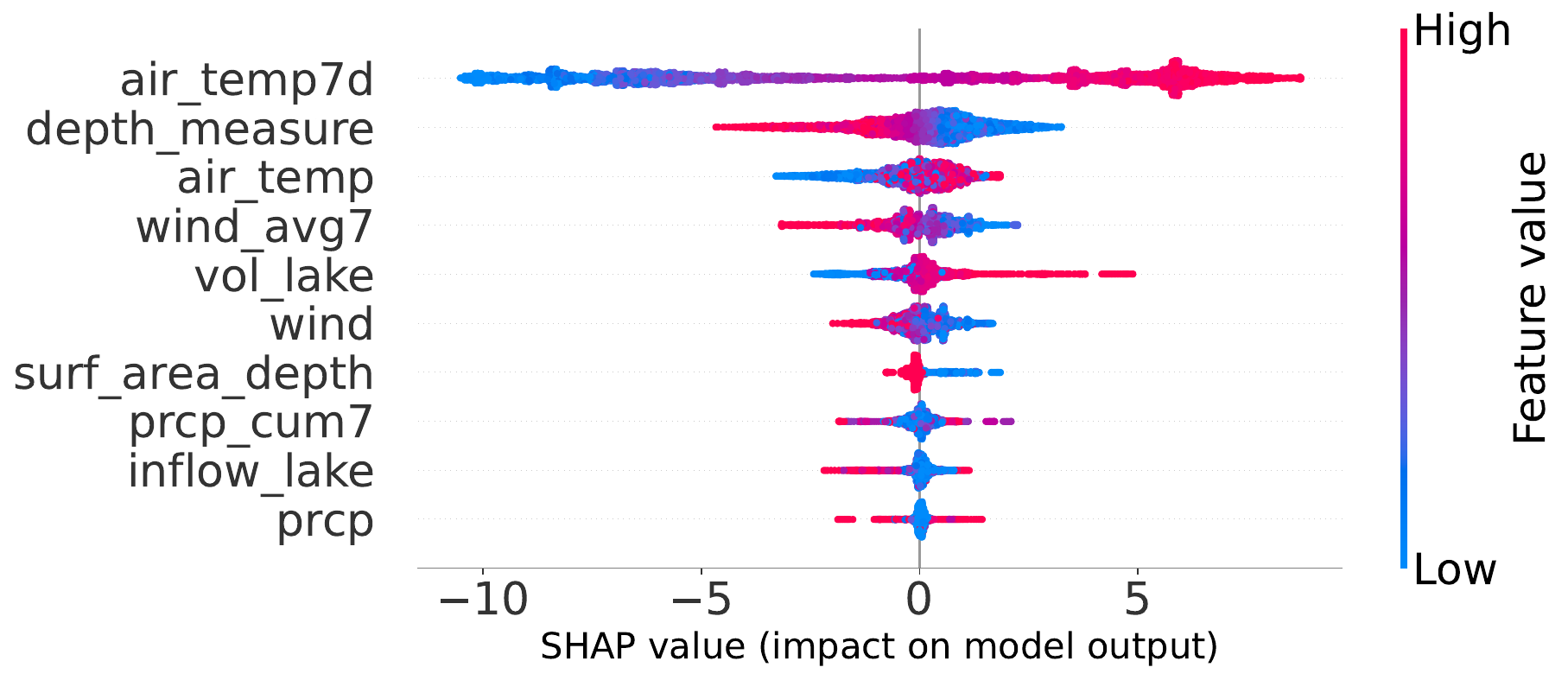}}\\
  \caption{SHAP summary plots of feature importance for the (a) RF, (b) XGBoost, and (c) MLP models. The y-axis lists input variables ranked by their overall importance, from most to least influential. The x-axis shows the distribution of SHAP values across all test samples, representing the magnitude and direction of each feature's impact on the model output. Each point corresponds to a single prediction; its position on the x-axis indicates the SHAP value (i.e., the feature’s contribution to the predicted reservoir water temperature, RWT), while its color reflects the original value of the feature (red for high values, blue for low values). Positive SHAP values indicate that the feature increased the predicted RWT, while negative values indicate a decreasing effect.}
  \label{summarizeSHAP}
\end{figure}

\begin{figure}
  \centering
  \subfloat[RF]{\includegraphics[width=0.71\linewidth]{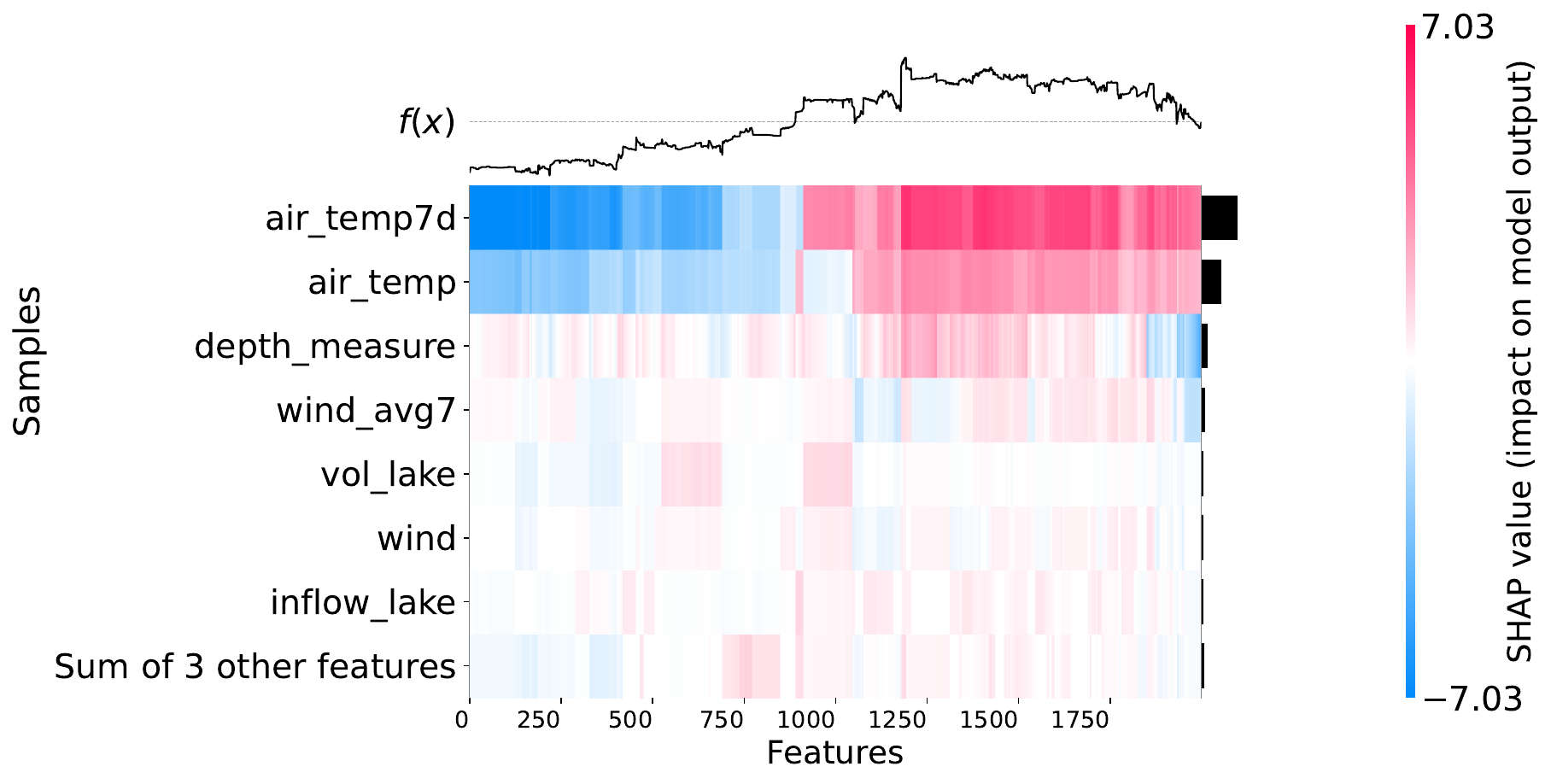}}\hfill 
  \subfloat[XGBoost]{\includegraphics[width=0.71\linewidth]{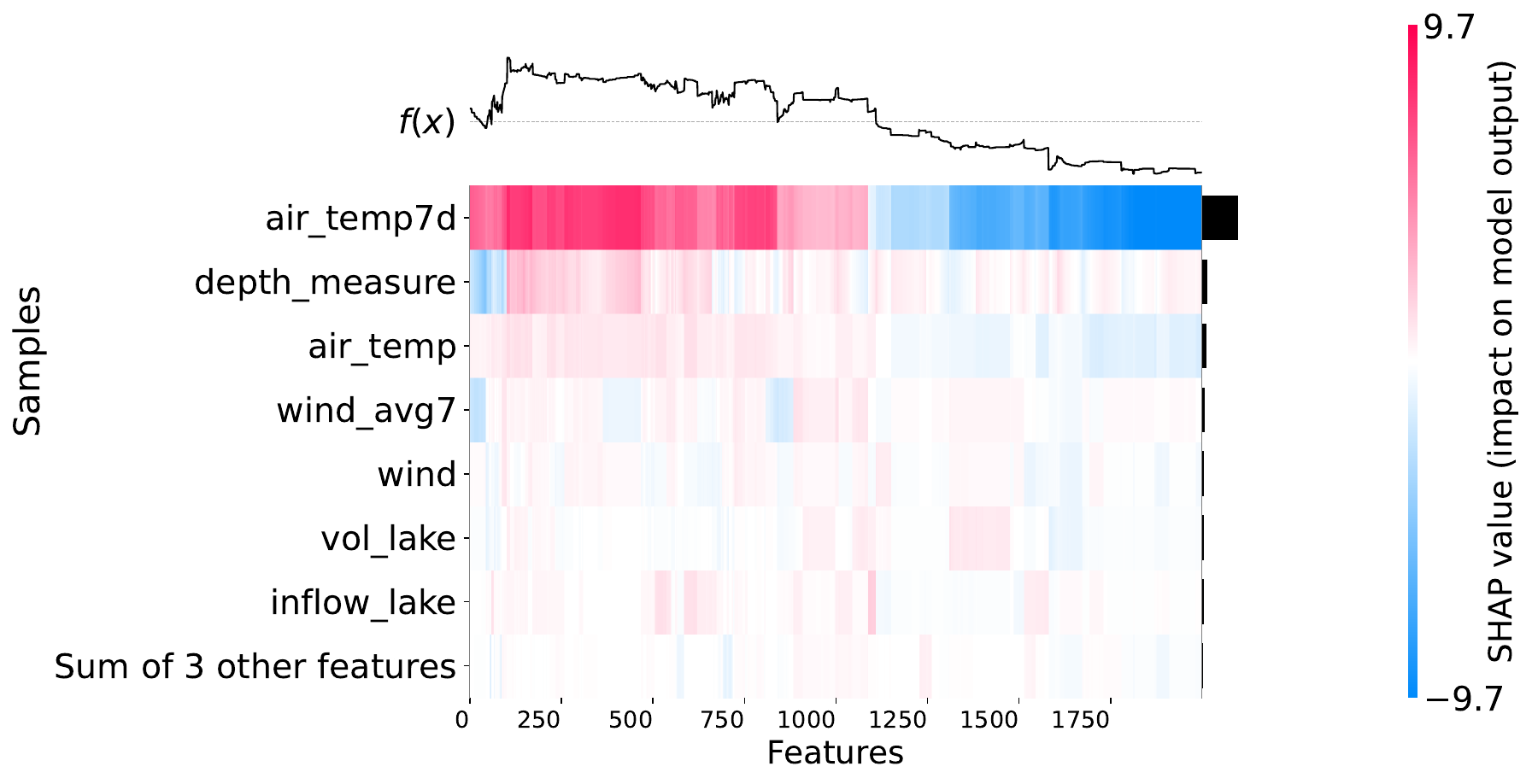}}\hfill
  \subfloat[MLP]{\includegraphics[width=0.71\linewidth]{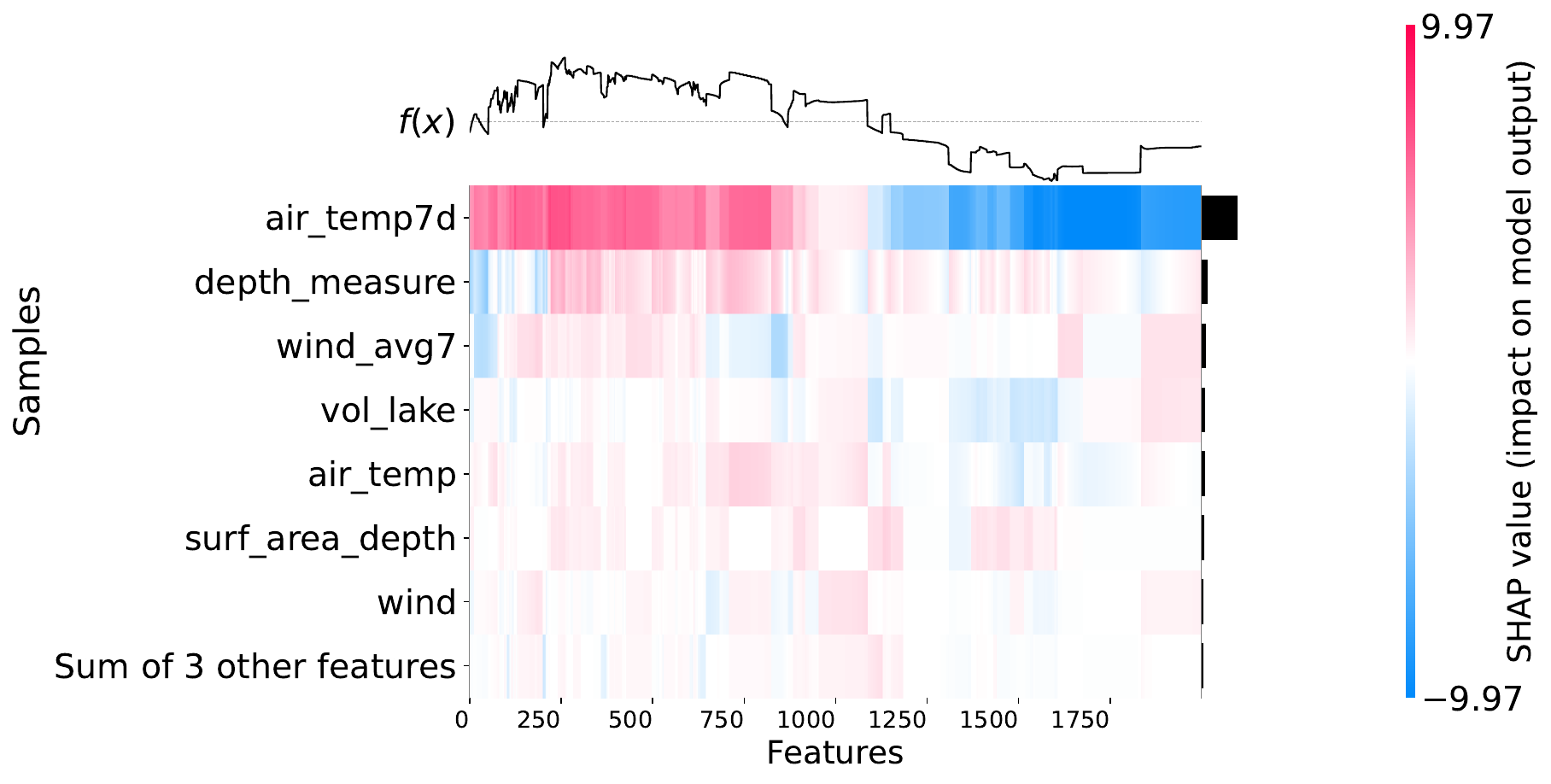}}\\
  \caption{Heatmap plots of SHAP values for all input features across all samples in the test set for the RF, XGBoost, and MLP models. Each column corresponds to a specific sample (e.g., a reservoir-day pair), and each row represents an input predictor. Cell colors indicate the SHAP value for that feature and sample: red values signify a positive contribution to the predicted output f(x), while blue values indicate a negative contribution, with darker shades representing greater magnitude. f(x) denotes the model’s prediction for each sample, shown alongside the rows. The width of the black bar on the right-hand side of each column reflects the global importance of the corresponding feature. }
  \label{heatmap}
\end{figure}

Figure \ref{summarizeSHAP} shows SHAP summary plots illustrating the impact of input features on RWT predictions for the RF, XGBoost, and MLP models. In comparison, Figure \ref{heatmap} presents SHAP heatmaps that depict the contributions of each input feature across all test samples for these models. Both plots help highlight the most impactful variables within each model. 

Consistently, \textit{air\_temp7d} stands out as the most important variable to predict RWT across models, always contributing between 46.7\% and 63.7\% to the models' predictions (see Fig.\ref{heatmap}(a) through (c) black side bars and Table \ref{shap_contributions}). \textit{air\_temp7d}'s SHAP values transition from negative at low feature values to positive at high values, indicating that cooler air temperatures lead to lower predicted RWT, while warmer air temperatures result in higher predictions (Fig.\ref{summarizeSHAP}). This pattern is visually reinforced by the color gradient from blue to red, where the intensity of the SHAP values reflects the strength of each feature's contribution to the model (Fig.\ref{summarizeSHAP}(a) through (c)). For the rest of the predictors' importance, there are observable differences across models as described in the following paragraphs, but in general, patterns remain consistent for a subset of variables. Features such as \textit{depth\_measure}, \textit{air\_temp}, and \textit{wind\_avg7} consistently rank among the top contributors, although their relative influence and direction of impact vary by model architecture.

In the RF model (Table \ref{shap_contributions}, Figs.\ref{summarizeSHAP}(a) and \ref{heatmap}(a)), \textit{air\_temp7d}, \textit{air\_temp}, and \textit{depth\_measure} emerge as the top predictors, contributing 46.66\%, 25.54\%, and 9.5\%, respectively, and collectively explaining 81.7\% of the variability in RWT. Both temperature-related features follow the expected pattern: low values contribute to reduced RWT predictions, while high values increase them (Fig.\ref{summarizeSHAP}(a)). On the other hand, \textit{depth\_measure} shows a negative relationship with greater depths resulting in cooler predicted temperatures, likely due to thermal stratification. This trend is consistent across both summary plots and the SHAP heatmap (Fig.\ref{summarizeSHAP}(a) and \ref{heatmap}(a)), where negative SHAP values dominate for deep observations. The heatmap in Fig.\ref{heatmap}(a) also shows a more gradual shift in SHAP contributions compared to MLP (Fig.\ref{heatmap}(b)) and XGBoost (Fig.\ref{heatmap}(c)), reflecting the RF model’s ensemble structure and less abrupt response to feature interactions.

The XGBoost model (Table \ref{shap_contributions}, Figs.~\ref{summarizeSHAP}(b) and~\ref{heatmap}(b)) presents a similar feature hierarchy, with \textit{air\_temp7d} again ranking first (63.74\% contribution), followed by \textit{depth\_measure} (9.94\% contribution), \textit{air\_temp} (8.31\% contribution), and \textit{wind\_avg7} (4.78\% contribution). A notable difference is the appearance of \textit{wind} among the top five predictors, replacing \textit{vol\_lake}, which was more relevant in the RF model. Summary plots show that increasing \textit{wind\_avg7} and \textit{wind} reduce predicted RWT, consistent with enhanced cooling through surface mixing. The heatmap in Fig.\ref{heatmap}(b) confirms these patterns, where instances with stronger wind conditions exhibit predominantly negative SHAP values for these variables. Additionally, the hierarchical clustering (Fig.\ref{summarizeSHAP}(b))  reveals groupings of instances with shared patterns, such as low wind and shallow depth leading to higher predictions.

In the MLP model (Table \ref{shap_contributions}, Figs.~\ref{summarizeSHAP}(c) and~\ref{heatmap}(c)), the top-ranked variables are \textit{air\_temp7d}, \textit{depth\_measure}, \textit{air\_temp}, \textit{wind\_avg7}, and \textit{wind}, similar to XGBoost. As in the other models, higher values of \textit{air\_temp7d} strongly drive predictions upward, while greater depths and wind speeds reduce predicted RWT. Notably, the MLP heatmap (Fig.~\ref{heatmap}c) shows sharper transitions in SHAP values across clustered instances, indicating that the neural network captures nonlinear relationships and interactions more distinctly than tree-based models. One additional insight from the MLP model is the contribution of features like \textit{surf\_area\_depth}, which, although not among the top five, exhibits strong positive SHAP values at lower feature values, potentially indicating that shallow lakes with large surface areas are more sensitive to atmospheric heating.

While the dominant predictors are broadly consistent across all models, differences in ranking and variable sensitivity, such as the higher importance of \textit{vol\_lake} in RF versus \textit{wind} in MLP and XGBoost, highlight how each architecture handles feature interactions. For instance, \textit{vol\_lake} contributes positively in RF, possibly due to increased thermal inertia in larger lakes, but appears less critical in MLP and XGBoost. Overall, integrating both summary and heatmap visualizations provides a comprehensive view of how features contribute to RWT predictions, revealing both consistent drivers and model-specific dynamics.

\begin{table}
\small
\centering
\caption{Global SHAP Feature Contribution Percentages (\%) across Models}
\label{shap_contributions}
\begin{tabularx}{\textwidth}{l *{3}{>{\centering\arraybackslash}X}}
\toprule
\textbf{Feature} & \textbf{RF (\%)} & \textbf{XGBoost (\%)} & \textbf{MLP (\%)} \\
\midrule
air\_temp7d         & 46.66 & 63.74 & 60.62 \\
air\_temp           & 25.54 & 8.31  & 6.04  \\
depth\_measure      & 9.50  & 9.94  & 9.31 \\
wind\_avg7          & 4.90  & 4.78  & 7.17  \\
vol\_lake           & 3.42  & 3.39  & 4.50 \\
inflow\_lake        & 3.07  & 3.79  & 2.31  \\
wind                & 2.80  & 3.72  & 4.15  \\
surf\_area\_depth   & 0.34  & 0.16  & 2.17  \\
prcp                & 2.05  & 0.94  & 1.06  \\
prcp\_cum7          & 1.73  & 1.24  & 2.61  \\

\bottomrule
\end{tabularx}
\end{table}

\vspace{0.5cm}

\subsection{KAN and Proposed Symbolic Equations}

Table~\ref{equations_10} summarizes the model performance ($R^2$) for both the \textit{simple} and \textit{complex} KAN models, as well as the symbolic equations derived for each input set in the \textit{simple} model set. Each model set incrementally incorporated input features based on their importance ranking derived from SHAP analysis of the RF model, which outperformed XGBoost and MLP models in preliminary evaluations. The inputs were added in the following order: 7-day average air temperature (air\_temp7d, \( x_1 \)), same-day air temperature (air\_temp, \( x_2 \)), depth (depth\_measure, \( x_3 \)), 7-day average wind speed (wind\_avg7, \( x_4 \)), reservoir volume (vol\_lake, \( x_5 \)), same-day wind speed (wind, \( x_6 \)), surface area to maximum depth ratio (surf\_area\_depth, \( x_7 \)), and same-day and 7-day antecedent cumulative precipitation (prcp, \( x_9 \) and prcp\_cum7, \( x_{10} \)).

While the full symbolic equations of the \textit{complex} models are provided in Appendix A, their structure and performance characteristics, along with those of the \textit{simple} models, are further examined in the following sections.

The \textit{simple} model set was explicitly designed to maximize interpretability, typically comprising a combination of linear terms and rational (inverse) expressions. These forms enable a linear understanding of how each input contributes to the predicted output, which is particularly valuable for scientific interpretation. In this model set, the first model, which used only air\_temp7d (\( x_1 \)), yielded an $R^2$ of 0.8207. With each additional variable, performance increased incrementally; however, the magnitude of performance gain was not uniform. The largest single-step improvement occurred when depth (\( x_3 \)) was added as the third input, increasing $R^2$ from 0.8224 to 0.8441. This suggests that depth plays a significant role in explaining RWT variability in combination with air temperature. In contrast, the smallest performance gain occurred after adding the final input, same-day precipitation (\( x_{10} \)), which improved $R^2$ only marginally from 0.8801 to 0.8805. This indicates diminishing returns beyond the inclusion of the top seven to eight variables.

In this \textit{simple} model set, the linear coefficients across models for several variables remain consistent in magnitude, particularly for air\_temp7d (\( x_1 \)) (between 0.81 and 0.86) and depth (\( x_3 \)) (between 0.14 and 0.15), suggesting a stable and direct influence of these variables on RWT. As more inputs are added, the KAN equations begin to incorporate nonlinear terms (e.g., rational or squared inverse terms) for lower-ranked variables like vol\_lake (\( x_5 \)), prcp\_cum7 (\( x_9 \)), and prcp (\( x_{10} \)). Importantly, even when such nonlinear terms are introduced, the symbolic forms remain tractable, reinforcing the interpretability advantage of KAN over traditional black-box neural networks.

\renewcommand{\arraystretch}{1.3} 
\setlength{\arrayrulewidth}{0.2pt} 

\begin{longtable}{
    |>{\centering\arraybackslash}m{3.8cm}
    |>{\centering\arraybackslash}m{2.2cm}
    |>{\centering\arraybackslash}m{2.2cm}
    |>{\centering\arraybackslash}m{\dimexpr\textwidth - 8.7cm - 3\tabcolsep - 4\arrayrulewidth}|
}

\caption{KAN models summary with inputs, performance, and derived symbolic equations. Each KAN model was trained using an incremental set of input variables selected according to their importance ranking from Random Forest-based SHAP analysis. All variables were normalized to the range [0, 1] using Min-Max scaling. The table presents, for each model, the specific input set used, the corresponding $R^2$ value on the test set for both the interpretable (\textit{simple}) model and a more complex version (with higher accuracy but lower interpretability), and the symbolic equation form of the \textit{simple} model within the normalized feature space. The more complex equations can be found in the Appendix A. These equations are valid only within the training input range (see Table \ref{TableFeatures}).}
\label{equations_10} \\
\hline
\textbf{Inputs} & \textbf{Complex Model (R\(^2\))} & \textbf{Simple Model (R\(^2\))} & \textbf{Equation (Simple)} \\
\hline
\endfirsthead

\hline
\textbf{Inputs} & \textbf{Complex Model (R\(^2\))} & \textbf{Simple Model (R\(^2\))} & \textbf{Equation (Simple)} \\
\hline
\endhead

\hline
\multicolumn{4}{r}{\small\itshape Continued on next page} \\
\endfoot

\hline
\endlastfoot


\begin{minipage}{\linewidth}
\centering
\small
\textit{}\\
air\_temp7d (\(x_1\)) \\
\textit{}
\end{minipage}
&
\small
0.8421
&
\small
0.8207
&
\begin{minipage}{\linewidth}
\begingroup
\footnotesize
\begin{equation}
\begin{aligned}
y &= 0.85\,x_{1} + 0.04
\end{aligned}
\end{equation}
\endgroup
\end{minipage}
\\
\hline

\begin{minipage}{\linewidth}
\centering
\small
\textit{}\\
air\_temp7d (\(x_1\)) \\
air\_temp (\(x_2\))\\
\textit{}
\end{minipage}
&
\small
0.8474
&
\small
0.8224
&
\begin{minipage}{\linewidth}
\begingroup
\footnotesize
\begin{equation}
\begin{aligned}
y &= 0.85\,x_{1} + \frac{0.013}{-4.8 x_2 - 0.19}
+ 0.05
\end{aligned}
\end{equation}\
\endgroup
\end{minipage}
\\
\hline

\begin{minipage}{\linewidth}
\centering
\small
\textit{}\\
air\_temp7d (\(x_1\)) \\
air\_temp (\(x_2\))\\
depth\_measure (\(x_3\))\\
\textit{}
\end{minipage}
&
\small
0.8773
&
\small
0.8441
&
\begin{minipage}{\linewidth}
\begingroup
\footnotesize
\begin{equation}
\begin{aligned}
y &= 0.87\,x_{1} - \frac{0.009}{- 9.7 x_2 +9.8}
- 0.15\,x_{3}
+ 0.06
\end{aligned}
\end{equation}
\endgroup
\end{minipage}
\\
\hline

\begin{minipage}{\linewidth}
\centering
\small
\textit{}\\
air\_temp7d (\(x_1\)) \\
air\_temp (\(x_2\))\\
depth\_measure (\(x_3\))\\
wind\_avg7 (\(x_4\))\\
\textit{}
\end{minipage}
&
\small
0.8859
&
\small
0.8611
&
\begin{minipage}{\linewidth}
\begingroup
\footnotesize
\begin{equation}
\begin{aligned}
y &= 0.82\,x_{1} 
+ \frac{0.012}{- 0.2 x_2 -0.106}
- 0.15\,x_{3}\\
&\quad - 0.10\,x_{4}
+ 0.235
\end{aligned}
\end{equation}
\endgroup
\end{minipage}
\\
\hline

\begin{minipage}{\linewidth}
\centering
\small
\textit{}\\
air\_temp7d (\(x_1\)) \\
air\_temp (\(x_2\))\\
depth\_measure (\(x_3\))\\
wind\_avg7 (\(x_4\))\\
vol\_lake (\(x_5\))\\
\textit{}
\end{minipage}
&
\small
0.8982
&
\small
0.8639
&
\begin{minipage}{\linewidth}
\begingroup
\footnotesize
\begin{equation}
\begin{aligned}
y &= 0.85\,x_{1} 
+ \frac{0.0026}{- 3.0 x_2 -0.092}
- 0.14\,x_{3}\\
&\quad - 0.15\,x_{4}
 + \frac{0.0004}{- 7.2 x_5 +3.5}
+ 0.1528
\end{aligned}
\end{equation}
\endgroup
\end{minipage}
\\
\hline

\begin{minipage}{\linewidth}
\centering
\small
\textit{}\\
air\_temp7d (\(x_1\)) \\
air\_temp (\(x_2\))\\
depth\_measure (\(x_3\))\\
wind\_avg7 (\(x_4\))\\
vol\_lake (\(x_5\))\\
wind (\(x_6\))\\
\textit{}
\end{minipage}
&
\small
0.9053
&
\small
0.8701
&
\begin{minipage}{\linewidth}
\begingroup
\footnotesize
\begin{equation}
\begin{aligned}
y &= 0.81\,x_{1} 
+ \frac{0.0079}{- 0.6 x_2 -0.04}
- 0.14\,x_{3}\\
&\quad - 0.15\,x_{4}  + \frac{0.0003}{- 0.6 x_5 +0.032}\\
&\quad - 0.0806\,x_{6}
+ 0.2224
\end{aligned}
\end{equation}
\endgroup
\end{minipage}
\\
\hline

\begin{minipage}{\linewidth}
\centering
\small
\textit{}\\
air\_temp7d (\(x_1\)) \\
air\_temp (\(x_2\))\\
depth\_measure (\(x_3\))\\
wind\_avg7 (\(x_4\))\\
vol\_lake (\(x_5\))\\
wind (\(x_6\))\\
surf\_area\_depth (\(x_7\))\\
\textit{}
\end{minipage}
&
\small
0.9115
&
\small
0.8750
&
\begin{minipage}{\linewidth}
\begingroup
\footnotesize
\begin{equation}
\begin{aligned}
y &= 0.81\,x_{1} 
+ \frac{0.003}{- 0.002 x_2 -0.008}
- 0.14\,x_{3}\\
&\quad
- 0.15\,x_{4} - \frac{0.008}{- 10.0 x_5 -0.2} \\
&\quad - 0.081\,x_{6}
+ 0.009\,x_{7}
+ 0.556
\end{aligned}
\end{equation}
\endgroup
\end{minipage}
\\
\hline

\begin{minipage}{\linewidth}
\centering
\small
\textit{}\\
air\_temp7d (\(x_1\)) \\
air\_temp (\(x_2\))\\
depth\_measure (\(x_3\))\\
wind\_avg7 (\(x_4\))\\
vol\_lake (\(x_5\))\\
wind (\(x_6\))\\
surf\_area\_depth (\(x_7\))\\
inflow\_lake (\(x_8\))\\
\textit{}
\end{minipage}
&
\small
0.9130
&
\small
0.8795
&
\begin{minipage}{\linewidth}
\begingroup
\footnotesize
\begin{equation}
\begin{aligned}
y &= 0.83\,x_{1} 
+ \frac{0.018}{-1.40\,x_{2} - 0.18} 
- 0.15\,x_{3} \\
&\quad
- 0.14\,x_{4}  + \frac{0.002}{0.18 - 1.41\,x_{5}}
- 0.074\,x_{6} \\
&\quad
- 0.016\,x_{7} 
+ 0.012\,x_{8} 
+ 0.21
\end{aligned}
\end{equation}
\endgroup
\end{minipage}
\\
\hline

\begin{minipage}{\linewidth}
\centering
\small
\textit{}\\
air\_temp7d (\(x_1\)) \\
air\_temp (\(x_2\))\\
depth\_measure (\(x_3\))\\
wind\_avg7 (\(x_4\))\\
vol\_lake (\(x_5\))\\
wind (\(x_6\))\\
surf\_area\_depth (\(x_7\))\\
inflow\_lake (\(x_8\))\\
prcp\_cum7 (\(x_9\))\\
\textit{}
\end{minipage}
&
\small
0.9182
&
\small
0.8801
&
\begin{minipage}{\linewidth}
\begingroup
\footnotesize
\begin{equation}
\begin{aligned}
y &= 0.84\,x_{1} 
+ \frac{0.0.15}{-2.81\,x_{2} - 0.19} 
- 0.15\,x_{3} \\
&\quad
- 0.15\,x_{4}  + \frac{0.00012}{ - 0.99\,x_{5} +0.50}
- 0.071\,x_{6} \\
&\quad
- 0.04\,x_{7}   - 0.015\,x_{8}\\
&\quad
+ \frac{0.00012}{(0.80 - x_{9})^{2}} 
+ 0.20 
\end{aligned}
\end{equation}
\endgroup
\end{minipage}
\\
\hline

\begin{minipage}{\linewidth}
\centering
\small
\textit{}\\
air\_temp7d (\(x_1\)) \\
air\_temp (\(x_2\))\\
depth\_measure (\(x_3\))\\
wind\_avg7 (\(x_4\))\\
vol\_lake (\(x_5\))\\
wind (\(x_6\))\\
surf\_area\_depth (\(x_7\))\\
inflow\_lake (\(x_8\))\\
prcp\_cum7 (\(x_9\))\\
prcp (\(x_{10}\))\\
\textit{}
\end{minipage}
&
\small
0.9220
&
\small
0.8805
&
\begin{minipage}{\linewidth}
\begingroup
\footnotesize
\begin{equation}
\begin{aligned}
y &= 0.86\,x_{1} 
+ \frac{0.011}{-5.0\,x_{2} - 0.19} 
- 0.14\,x_{3} \\
&\quad - 0.14\,x_{4}\ - \frac{0.004}{-0.39\,x_{5} - 0.14} 
- 0.071\,x_{6} \\
&\quad 
- 0.012\,x_{7} 
+ 0.063\,x_{8}\\
&\quad  + \frac{0.12}{(-0.18x_{9}-1 )^{2}} 
 \ 
\\
&\quad  + \frac{1.0 \times 10^{-5}}{(-x_{10} - 0.02)^{2}} 
 + 0.03
\end{aligned}
\end{equation}
\endgroup
\end{minipage}
\\
\hline

\end{longtable}

In contrast, the \textit{complex} models were constructed to optimize predictive accuracy without constraints on interpretability. As a result, their symbolic equations include highly nonlinear transformations, such as nested trigonometric, exponential, and polynomial terms, often involving multiple interacting variables (see Appendix A). For instance, while a simple model might include additive linear contributions from normalized features, a complex model with the same inputs may embed intricate expressions such as $\cos(ax + b)$, $e^{cx}$, or $\tan(dx + e)$ within a composite rational function. These richer functional forms capture subtle, high-order dependencies in the data, resulting in improved predictive performance. For example, using all ten inputs, the \textit{simple} model achieved an $R^2$ of 0.8805, whereas the corresponding \textit{complex} model reached an $R^2$ of 0.9220, representing a relative improvement of approximately 4.7\%. However, this gain comes at the cost of reduced linear interpretability and potentially lower generalizability outside the training data domain.

\begin{figure}
	\centering
		\includegraphics[width=\textwidth]{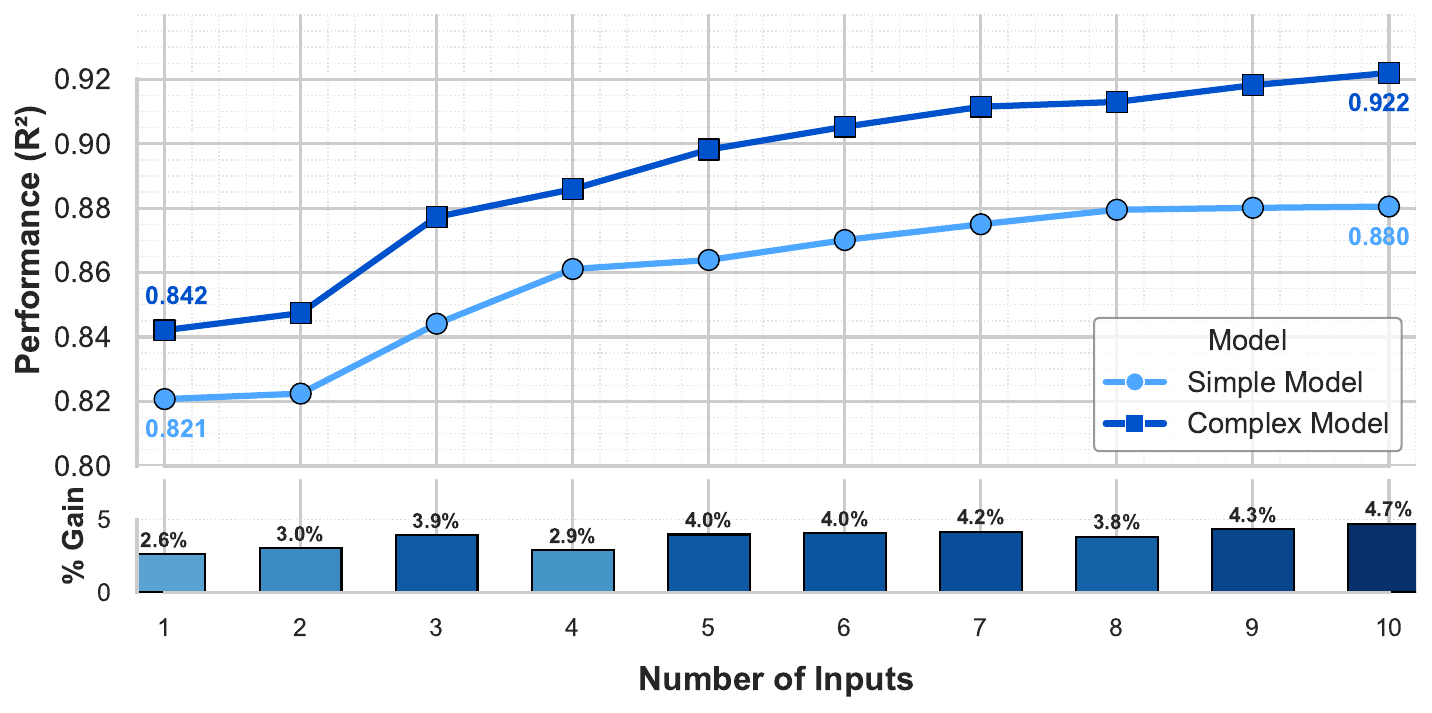}
	\caption{Model performance (R\textsuperscript{2}) of the KAN models as a function of the number of input features used to predict water temperature. Input variables were sequentially added based on their ranked importance derived from feature importance analysis using the RF model, which outperformed both XGBoost and MLP models in preliminary evaluations. The plot compares two versions of the KAN model: a \textit{simple} model (solid line) and a \textit{complex} model (dashed line). The \textit{simple} model's performance increased from R\textsuperscript{2} = 0.8207 with a single input (air\_temp7d) to R\textsuperscript{2} = 0.8805 when all ten inputs were included. The \textit{complex} model consistently outperformed the \textit{simple} model, achieving an R\textsuperscript{2} of 0.9220 with all ten inputs, demonstrating the benefits of increased model complexity.}
	\label{RvsP}
\end{figure}

To illustrate the trade-off between model complexity and performance, Figure~\ref{RvsP} shows the $R^2$ values for both \textit{simple} and \textit{complex} models trained on the same input sets. While the complex models typically achieve higher predictive accuracy due to their greater representational capacity, the simpler models produce closed-form equations that are easier to interpret and communicate. Our focus is on these interpretable symbolic models, which offer a practical balance between accuracy and transparency, an essential consideration for stakeholders in data-driven decision-making.

\begin{figure}
  \centering
  \subfloat[Notations of activations that flow through the KAN network.]{\includegraphics[width=0.48\linewidth]{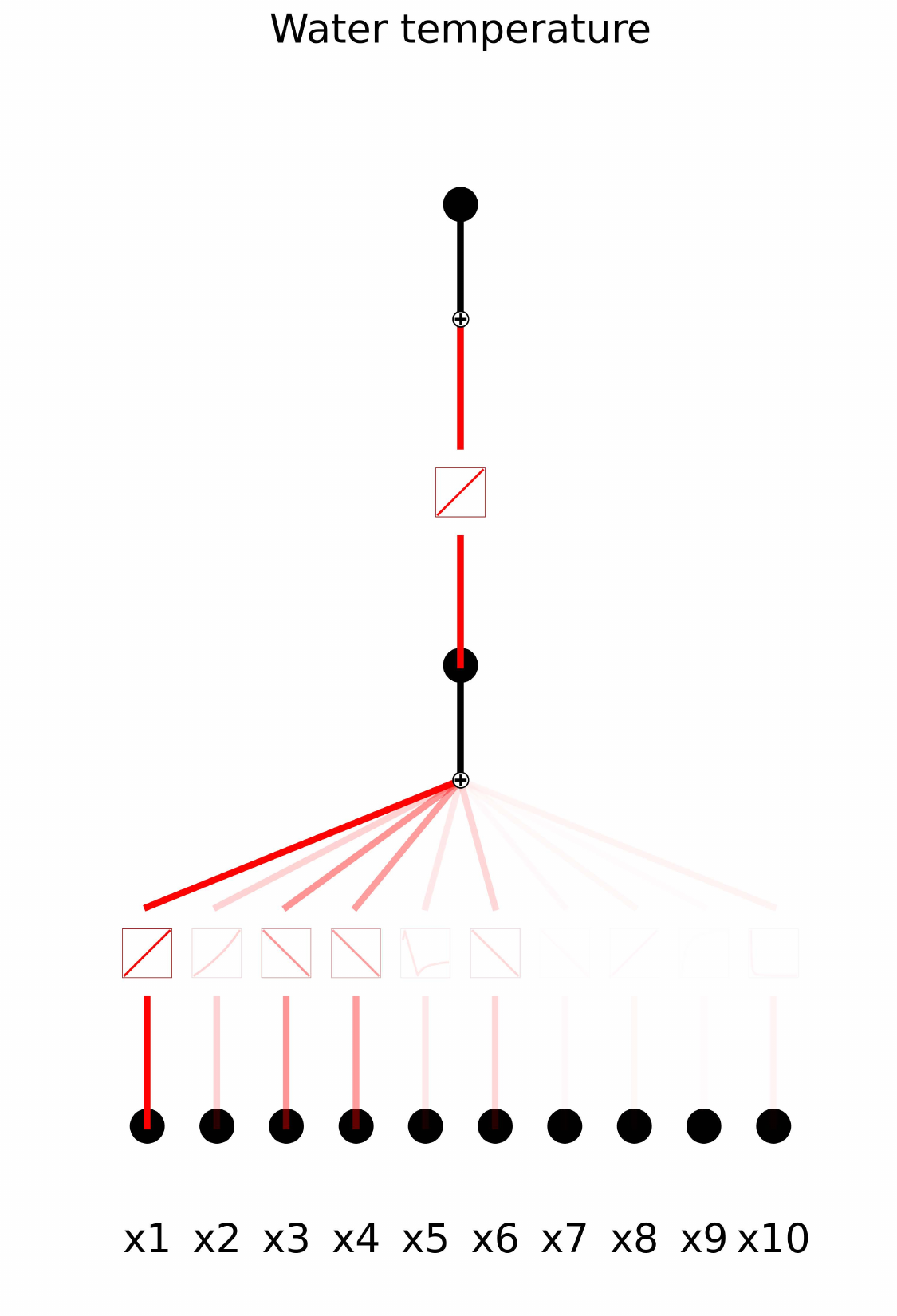}}\hfill 
  \subfloat[Performance of the KAN model for reservoir water temperature after hyperparameter optimization.]{\includegraphics[width=0.48\linewidth]{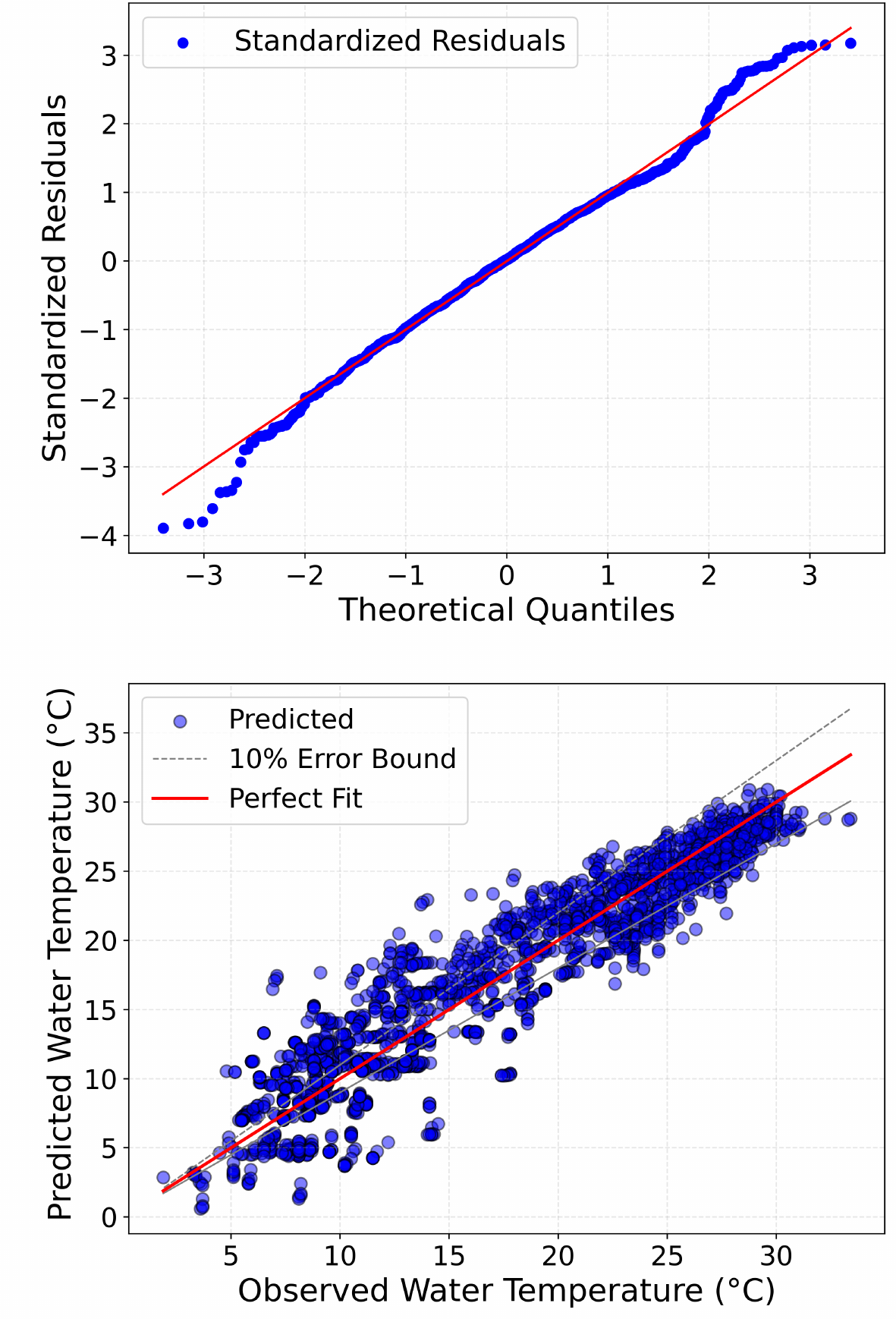}}\\
  \caption{KAN model results for the \textit{simple} model using 10 input variables. (a) Network with activation functions on the edges. The network includes ten input features and one output node. Darker red edges indicate stronger contributions of individual input features, with each edge representing a distinct parameterized univariate activation function. These outputs are then passed through one additional node, with the resulting activations summed to produce the features for the output node. (b) KAN model’s performance, where (Upper) is the quantile plot of residuals, and (lower) displays the measured values (x-axis) versus the predicted values (y-axis).}
  \label{kanGraph}
\end{figure}

Figure \ref{kanGraph} illustrates the performance of the calibrated KAN model for predicting water temperature during the test phase using all ten input variables from the \textit{simple} model set. Fig.\ref{kanGraph}(a), shows the ten input variables and their relationships are shown in the network, with darker connections indicating a higher significance level. Additionally, Fig.\ref{kanGraph}(b, top), presents a quantile plot, where the data residuals align with the expected distribution, and the scatter plot (bottom), where the simulated data is centered around the expected value. However, some data points fall outside the 10\% boundary. The coefficient of determination ($R^2$) achieved during the testing phase was 0.8805. 

\section{Discussion}
\subsection{Context and summary of key findings}
Traditional approaches to modeling RWT, whether physics-based or empirical, often face challenges such as site-specific calibration demands, computational intensity, and limited scalability \citep{polli2019}. Recent advances in machine learning offer promising alternatives by capturing complex, nonlinear interactions among drivers across spatial and temporal scales 
\citep{sheik2024machine,liu2022permeability}. However, most studies still emphasize predictive accuracy at the expense of interpretability \citep{eshetu2023interpretable}, limiting process understanding and broader application.

In this study, we addressed these gaps by developing, training and testing a set of ML models, RF, XGBoost, and MLP and KAN to predict RWT across ten reservoirs in the climatically and morphometrically diverse Red River Basin of the south-central United States. Our modeling leveraged over 10,000 depth-resolved temperature measurements collected from 1996 to 2020, incorporating multiscale environmental predictors such as air temperature, wind speed, precipitation, lake volume, measurement depth and morphometric ratios. Incorporating these predictors helps in capturing the complexity of thermal profiles more accurately.

RF, XGBoost and MLP demonstrated strong predictive skill, but RF achieved the highest overall performance ($R^2$ = 0.974, RMSE = 1.830°C), aligning with previous successes of RF in hydrothermal studies \citep{w11050910}. Complementing these black-box models, the KAN approach provided interpretable symbolic equations, offering a novel framework for interpretable and transferable RWT prediction. Collectively, these findings demonstrate the feasibility of using interpretable ML approaches to predict vertical thermal profiles in reservoirs, advancing both predictive capability and process insight critical for water resource management under climate uncertainty.

An examination of model performance across individual reservoirs (Table~\ref{ResultsTEST_ap}) further highlights both robustness and potential limitations. Across the ten reservoirs, the three classical ML models generally achieved high predictive skill, with $R^2$ values frequently exceeding 0.90 and RMSE values typically below 1.5°C. The RF model consistently delivered strong performance across most sites, including Fort Cobb, Foss, Hugo, Texoma, Pine Creek, and Tom Steed. XGBoost occasionally outperformed RF at individual reservoirs, such as Arbuckle, Sardis, and Waurika. Nonetheless, the high performance at Sardis should be interpreted with caution: only two profiles were available, one for training and one for testing. In such cases, model evaluation becomes highly sensitive to the characteristics of a single test profile, which may not reflect broader patterns. Given RF’s consistent performance elsewhere, the lower score at Sardis likely reflects the differences in architectures. The RF model tends to generalize better across the basin when enough data is available \citep{10.5555/2188385.2343682}; meanwhile, XGBoost and MLP may overfit more to the local behavior of each lake, which affects them less with the lack of more data \citep{XGBoost}. That situation is just a reflection of how architectures can use the same data differently \citep{10.1145/2347736.2347755}, and therefore, despite RF being the overall best model, we cannot consider that this situation will always be the case.

\subsection{Interpretation of model behavior and feature contributions}

A consistent pattern across the RF, XGBoost, and MLP models was the importance of air temperature and measurement depth as primary drivers of RWT. In particular, the 7-day antecedent average air temperature (air\_temp7d) emerged as the most influential predictor based on SHAP analyses, which reflects the physical processes governing heat exchange between the atmosphere and reservoir systems. The strong influence of the air\_temp7d is consistent with the thermal memory of large water bodies, where RWT does not respond instantaneously to atmospheric fluctuations but integrates heat inputs over time \citep{WETZEL2001151}. This multi-day average captures both short-term variability and the cumulative effects of sustained warming or cooling events, which are particularly relevant in systems with high thermal inertia. By smoothing transient fluctuations and emphasizing persistent weather patterns, air\_temp7d serves as a physically meaningful proxy for the net atmospheric energy input driving reservoir thermal dynamics, and reflects the memory of the system. Prior studies similarly emphasize the importance of lagged meteorological inputs in capturing RWT dynamics \citep{Piccolroaz, SCHMID2022467}.

Feature importance analyses, along with the form of the simplified KAN equations, indicated that depth exhibited a consistent negative relationship with RWT across all models. Deeper layers were associated with lower temperatures, an expected outcome given the attenuation of solar radiation with increasing depth and the stabilizing influence of thermal stratification during the summer months. In the dataset, 32\% of all profiles demonstrated stratified conditions. Major reservoirs such as Lake Texoma, Arbuckle Reservoir, and Sardis Lake exhibit well-defined summer stratification, with thermoclines typically developing at depths between 5 and 13 meters \citep{owrbLakesReport2007}. These patterns highlight how deeper water masses remain insulated from short-term atmospheric forcing, preserving cooler temperatures throughout the stratified period.

In the \textit{simple} KAN models, this behavior was captured by a linear negative coefficient (e.g., depth coefficient between \(-0.14\) and \(-0.15\)). While this linear approximation aligns with the overall trend of decreasing temperature with depth, it may not fully capture the nonlinear structure typically associated with thermal stratification, such as sharp inflections near the thermocline. However, given the depth range and resolution of the available data, the linear form provides a parsimonious and interpretable approximation that still reflects the dominant physical behavior.

Wind variables, particularly the 7-day average wind speed (wind\_avg7), also played a secondary, but visible, role generally contributing to reduced RWT predictions. This behavior is consistent with established understanding that stronger winds enhance surface mixing and promote heat loss, thereby lowering surface and near-surface temperatures \citep{woolway2017amplified,Piccolroaz}. The explicit inclusion of both short-term and antecedent wind metrics in our models helped capture the cumulative impact of wind-driven mixing events on reservoir thermal regimes.  In the \textit{simple} KAN models, this relationship was consistently captured through linear negative coefficients for both wind\_avg7 (\(x_4\)) and same-day wind speed (\(x_6\)). This direct linear relationship reinforces the physical interpretation of wind as a driver of surface heat loss and stratification disruption, mixing the water column and decreasing surface water temperature \citep{WETZEL2001151,imberger1989physical}.

\subsection{Interpretability and Knowledge Compression through Symbolic Modeling}

Unlike black-box models, which may require hundreds or thousands of parameters to approximate similar dynamics, KAN offers explicit equations that not only predict thermal profiles with reasonable fidelity but also reveal the relationships among input drivers in a transparent, physically interpretable algebraic form, particularly in the \textit{simple} KAN models.

For instance, the best-performing 10-input model in the \textit{simple} set achieved an $R^2$ of 0.8805 while maintaining a symbolic structure dominated by linear terms. The 7-day average air temperature consistently appeared as a strong positive linear predictor across all models, reinforcing its role as the primary driver of heat input \citep{Akazan2025}. In contrast, depth contributed negatively and linearly, reflecting the well-established decline in temperature with increasing depth due to stratification and the attenuation of solar radiation \citep{wang2023climate}. A few nonlinear terms, such as rational or inverse-square expressions involving secondary variables like precipitation and reservoir volume, likely captured feedback mechanisms, including threshold-like storm impacts on inflow-driven mixing or the thermal buffering capacity of large water volumes \citep{Ozaki1980}. Despite these nonlinearities, the resulting equations remained interpretable and avoided devolving into opaque or overfit forms, preserving the core advantage of symbolic regression.

In comparison, the \textit{complex} KAN models, built with deeper architectures and non-polynomial basis functions, achieved higher predictive performance (up to $R^2 = 0.922$) and stronger knowledge compression. These models distilled large volumes of nonlinear interactions into highly compact symbolic forms, but their reliance on nested trigonometric and exponential terms reduced interpretability and obscured connections to known physical processes. This underscores a central tradeoff: while more expressive architectures can yield superior accuracy and tighter compression, simpler models retain symbolic clarity that makes their insights more readily transferable to scientific understanding.

The combination of efficiency, predictive skill, and interpretability in our \textit{simple} KAN framework supports both accurate empirical modeling and improved process-based understanding. Models that explicitly encode physical principles, such as energy balance or depth-dependent stratification, are also better equipped to generalize beyond the training domain \citep{Kapoor2023}. This is especially critical in climate adaptation, where decision-makers must evaluate scenarios beyond historical precedent. Interpretable models like ours enable tracing of cause–and–effect relationships under future conditions, thereby enhancing confidence in scenario analysis and policy development. Moreover, in stakeholder-facing contexts, particularly those involving public resources such as water, transparent and explainable models foster trust, enable interdisciplinary communication, and facilitate the translation of scientific insights into actionable decisions.

A key innovation of our approach was the SHAP-informed ordering of inputs in the KAN experiments. By incrementally adding predictors based on their ranked contribution to RF model outputs (the top-performing model), we imposed a data-driven yet physically plausible structure to the feature inclusion process. Notably, the first three variables added, 7-day antecedent average air temperature, same-day air temperature, and depth, drove the largest single improvement in KAN performance, where $R^2$ increased from 0.8224 to 0.8441 when depth was added in the \textit{simple} model and from 0.8474 to 0.8773 in the \textit{complex} model, representing 2.6\% and 3.5\%. These variables are all grounded in thermal physics: air temperature controls the net atmospheric energy flux into the reservoir \citep{WETZEL2001151}; its 7-day average accounts for the system’s thermal inertia \citep{livingstone1998relationship}; and depth governs the vertical distribution of heat and the formation of stratified layers \citep{WETZEL2001151, imberger1989physical}.

Beyond the top five inputs, performance improvements followed a clear law of diminishing returns. The first five inputs collectively accounted for the majority of predictive gains, pushing $R^2$ to 0.864 in the \textit{simple} model and 0.8982 in the \textit{complex} model. Subsequent additions, such as lake morphometry and precipitation variables, introduced more complexity in the symbolic forms but yielded only marginal increases in predictive skill (e.g., less than 0.01 $R^2$ gain for the last three variables in the \textit{simple} model). This indicates that a limited set of key variables captures the majority of the explanatory power, allowing for parsimonious models without substantial loss in predictive accuracy.

By analyzing the inflection point at which complexity ceased to significantly improve performance, we identified a core subset of variables (air temperature, depth, wind speed, and reservoir volume) that capture the essential structure of reservoir thermodynamics in the Red River Basin. For stakeholders such as reservoir managers or monitoring agencies, this has practical value: fewer inputs reduce data collection costs, simplify model deployment, and make real-time or large-scale implementation more feasible. Moreover, the most influential inputs, such as air temperature, depth, wind speed, and volume, are either routinely monitored or available from public datasets, minimizing the need for specialized field instrumentation.

More broadly, our results suggest that future models can approach near-optimal accuracy using a reduced set of physically interpretable predictors. This is particularly valuable in regions or periods with sparse data, where the absence of high-resolution or continuous measurements limits the feasibility of more complex models. The ability to maintain predictive skill with minimal inputs enhances the model’s usability in data-limited settings and under budget constraints, supporting broader adoption of interpretable, process-consistent RWT predictions.

\subsection{Implications for water resource management and climate adaptation}

Overall, this study demonstrates the feasibility and value of using machine learning models, including RF, XGBoost, and MLP, as well as symbolic regression frameworks such as KAN, to predict behaviors, compress knowledge and interpret the drivers of reservoir water temperature. Achieving high predictive skill (R$^2$ = 0.92–0.97) with a minimal number of atmospheric, hydrological, and morphometric predictors highlights the models' capacity to capture essential thermal processes without requiring extensive or high-resolution datasets. This is particularly important for reservoirs where in situ monitoring is limited or intermittent, and where detailed process-based models are impractical to implement or calibrate. 

By relying on a parsimonious input structure, the modeling framework developed in this study offers a scientifically grounded alternative to data-intensive approaches, enabling first-order assessments of thermal dynamics in understudied or resource-constrained reservoir systems. Such capabilities are increasingly critical as climate change intensifies heatwaves, alters precipitation regimes, and disrupts stratification and mixing dynamics in freshwater bodies \citep{woolway2021phenological, wang2023climate}. In this context, our reduced-complexity framework provides actionable insights for water resource management, risk assessment, and climate adaptation. As water managers increasingly require reliable and transparent tools to anticipate thermal responses (because temperature directly influences water quality, ecosystem health, and regulatory compliance), our approach represents a practical “best available science” solution for data-limited environments.

Furthermore, the ability to link model predictions to interpretable equations not only enhances stakeholder trust and facilitates communication with policymakers, but also provides mechanistic insight into the causal drivers of reservoir thermal dynamics. By bridging predictive accuracy with physical interpretability and causal understanding, this work contributes to building decision-support capacity that enables more robust and environmentally responsive water management strategies. Such insights are essential for sustainably managing freshwater resources amid accelerating environmental change.

\subsection{Limitations}

While our models effectively predict RWT using a streamlined set of inputs, including air temperature, wind, precipitation, and key morphological and hydrological characteristics, some limitations warrant attention. In particular, the exclusion of certain secondary drivers such as water transparency and direct solar radiation may constrain predictive accuracy under specific conditions, especially in clear-water systems where radiative heat absorption fundamentally shapes thermal stratification \citep{henderson1988sensitivity,heiskanen2015effects,SST,Chakravarthy2022}. Also, salinity was not included. While salinity and temperature do not directly influence each other, both affect water density and thus jointly influence stratification and circulation \citep{10.1002/lno.10285}. 

However, the models have already achieved high performance metrics (e.g., $R^2$ > 0.94 in the best case - RF) without these inputs, suggesting that the core drivers capture the majority of the variability in the studied reservoirs. Moreover, variables such as salinity, solar radiation, and turbidity are not routinely measured across many monitoring programs, and their inclusion may introduce logistical and financial constraints, particularly in data-sparse regions. Therefore, while their integration could refine model performance in certain contexts, their omission reflects a deliberate trade-off between marginal accuracy gains and model generalizability, cost-efficiency, and scalability. Future work may prioritize these additions in targeted studies where optical dynamics are known to play a larger role or where high-resolution radiative data are already available.

Our study relied on data geographically and temporally constrained to reservoirs within the Red River Basin, reflecting specific climatic, hydrological, and infrastructural conditions. Consequently, our models are most applicable under constraints defined by the similarity of input conditions and system properties, such as comparable depth ranges, climatic regimes, and hydrological dynamics. Applying the model beyond these bounds could reduce accuracy and would likely require local calibration or retraining the models with new data. Expanding the dataset to cover a wider variety of reservoir types and climates would further strengthen model generalizability and robustness \citep{https://doi.org/10.1029/2018WR022643, zhu2020forecasting}.

\section{Conclusions}
This study trained and evaluated RF, XGBoost, MLP, and KAN models to predict reservoir water temperature (RWT) across ten climatically and morphometrically diverse reservoirs in the Red River Basin, USA. RF achieved the highest accuracy and was used to derive SHAP-based feature importance, which guided the development of symbolic KAN models. Key findings include

\begin{enumerate}
\item RF, XGBoost, and MLP predicted RWT with high accuracy (R$^2$ = 0.945–0.974; RMSE = 1.20–1.83 °C) using a concise and physically meaningful set of predictors that included antecedent air temperature, wind speed, precipitation, depth, and morphometric characteristics. SHAP analysis provided valuable cross-model insights into variable importance, though its interpretability remained limited for direct knowledge extraction.
\item KANs produced compact symbolic equations that balanced accuracy and interpretability. Simple KANs (R$^2$ up to 0.88) captured dominant linear and low-order nonlinear patterns consistent with known processes, while complex KANs (R$^2$ up to 0.92) improved accuracy through nested nonlinearities but reduced transparency, illustrating the trade-off between performance and clarity.
\item Simple KANs aligned well with physical understanding, particularly regarding sensitivity to air temperature and depth, reinforcing their credibility and usefulness for hypothesis testing, stakeholder communication, and applications requiring transparency or regulatory acceptance.
\item The proposed multi-level modeling framework supports the selection of models that best match the needs of different users, from interpretable and explanatory forms to performance-oriented predictive models, ensuring flexibility across scientific, regulatory, and management contexts.
\end{enumerate}

Overall, combining interpretable machine learning with environmental data enables robust and transparent modeling of reservoir thermal dynamics. Future work should incorporate additional drivers and broader spatial applications to strengthen adaptive water resource management under climatic variability.

\vspace{0.5cm}
\textit{Acknowledgments}

This research was supported by the U.S. Geological Survey – Climate Adaptation Science Centers (USGS‑CASC) grant No. 20006684, the National Science Foundation CAREER Award No. 2239550, the Army Research Office Grant No. W911NF‑24‑1‑0296 and the National Oceanic and Atmospheric Administration – Cooperative Science Center for Earth System Sciences and Remote Sensing Technologies under the Cooperative Agreement Grant NA22SEC4810016. 

The views and conclusions contained herein are solely those of the authors and do not necessarily reflect the views of the South Central Climate Adaptation Science Center, the U.S. Geological Survey, the National Science Foundation, the Army Research Office, the National Oceanic and Atmospheric Administration or the United States Government. Mention of trade names or commercial products does not constitute their endorsement by the U.S. Government. This manuscript is submitted for publication with the understanding that the United States Government is authorized to reproduce and distribute reprints for governmental purposes.

\newpage
\appendix
\section{KAN Complex Symbolic Equations}

\textbf{Complex 1-input (R\(^2\) = 0.8421)}
\begin{equation}
y = -0.36\cos(1.65 x_1 + 8.28) - 0.031 + \frac{2.0 \times 10^{-5}}{\left(-0.007 - e^{-9.99x_1}\right)^2}
\end{equation}


\textbf{Complex 2-input (R\(^2\) = 0.8474)
}
\begin{equation}
y = 1.85 - 1.76\cos\left( -0.29\cos(3.62x_1 + 6.16) + 13.19 + \frac{0.054}{-6.42x_2 - 0.18} \right)
\end{equation}


\textbf{Complex 3-input (R\(^2\) = 0.8773)
}
\begin{equation}
y = 1.64\left( e^{-2.33x_1} + 0.01\cos(10.0 x_2 - 4.6) + 0.15(-x_3 - 0.018)^2 - 0.75 \right)^2 + 0.17
\end{equation}



\textbf{Complex 4-input (R\(^2\) = 0.8859)
}
\begin{equation}
y = -0.122 - 
\frac{0.028}{
    -0.52e^{-4.93x_1}
    - 0.015\cos(4.75x_2 + 6.02)
    - 0.013\cos(2.22 x_3 - 3.21)
    - 0.02e^{2.18x_4}
    - 0.05
}
\end{equation}



\textbf{Complex 5-input (R\(^2\) = 0.8982)
}
\begin{equation}
\left.
y = 
\frac{0.061}{
    \begin{array}{l}
        - 0.86\tanh(3.32x_1 - 0.20) 
        - 0.01\cos(2.37x_2 - 7.70) 
        + 0.026\cos(2.36x_3 + 9.399) \\
        - 0.009\log(9.86 - 9.81x_4) 
        + 0.005\cos(8.96x_5 + 4.19) 
        + 0.99
    \end{array}
}
\right.
\end{equation}



\textbf{Complex 6-input (R\(^2\) = 0.9053)}
\begin{equation}
\left.
y = -2.091 +
\frac{0.075}{\begin{array}{l}
     1.61\, e^{-1.70(-x_1 - 0.84)^2}
    - 0.004\cos(7.20x_2 - 4.99)
    - 0.022\cos(2.26x_3 + 6.37) + 0.14\\
    - 0.041\cos(1.36x_4 + 6.02)
    - 0.004\cos(8.58x_5 + 1.01)
    - 0.015(0.83 - x_6)^2
    
\end{array}}
\right.
\end{equation}




\textbf{Complex 7-input (R\(^2\) = 0.9115)
}
\begin{equation}
\left.
y = -0.011 -
\frac{0.55}{\begin{array}{l}
    - 3.10\, e^{-4.67(-x_1 - 0.12)^2}
    - \dfrac{0.008}{0.18 - 2.42x_2} 
    - 0.68\, e^{-2.49(1 - 0.5x_3)^2} 
    - 0.79 \cos(0.88x_4 + 9.23) \\
    + 0.03 \cos(9.86x_5 + 0.79)
    + \dfrac{0.227}{(-0.28x_6 - 1)^2}
    + 0.011\, e^{1.6x_7}
    - 1.54 
\end{array}}
\right.
\end{equation}

\textbf{Complex 8-input (R\(^2\) = 0.9130)
}
\begin{equation}
\left.
y = -0.0037 -
\frac{0.026}{\begin{array}{l}
    - 0.247 e^{-3.17(-x_1 - 0.374)^2}
    - 0.0327 e^{-5.032x_2}
    - 0.0091(-x_3 - 0.34)^2 
    - 0.0002 e^{3.8x_4} \\ 
    + \dfrac{0.0003}{0.16 - 0.81x_5}
    - 0.0053x_6 
    - 0.0026\cos(5.0014x_7 - 0.19)\\
    + 0.018(0.49 - x_8)^2
    - 0.027

\end{array}}
\right.
\end{equation}

\textbf{Complex 9-input (R\(^2\) = 0.9182)
}
\begin{equation}
\left.
y = -0.30 -
\frac{0.013}{\begin{array}{l}
    0.02(1 - 0.96x_1)^2 
    - 0.001 e^{-11.8(0.063 - x_2)^2}
    - 0.002(-x_3 - 0.23)^2
    - 5.0 \times 10^{-5} e^{3.8x_4} \\
    + \dfrac{0.0003}{0.37 - 2.0x_5}
    + 0.004 e^{-0.46x_6} 
    + 0.001\cos(7.13x_7 + 2.81)\\
    + \dfrac{1.0\times10^{-5}}{(-x_8 - 0.092)^2} 
    + \phantom{+}0.0023(0.35 - x_9)^2 - 0.014
\end{array}}
\right.
\end{equation}

\textbf{Complex 10-input (R\(^2\) = 0.9220)
}
\begin{equation}
\left.
y = -0.42 +
\frac{0.062}{\begin{array}{l}
    0.036\cos(2.71x_1 + 0.72)
    + 0.011(0.63 - x_2)^2 
    + 0.007(-x_3 - 0.36)^2 + 0.004e^{3.2x_4} \\
     - 0.017e^{-10.0x_5}
    + 0.002\cos(3.54x_6 + 3.42) 
    - 0.009\cos(5.26x_7 + 9.36) + 0.10\\
     - 0.02(0.45 - x_8)^2
    -0.011(0.34 - x_9)^2 
    + 3.0 \times 10^{-5}\tan(9.40x_{10} + 1.60)
    
\end{array}}
\right.
\end{equation}

\bibliographystyle{model1-num-names}  

\bibliography{cas-refs}


\end{document}